\documentclass[twoside]{article}

\usepackage{pdfprivacy}
\usepackage[accepted]{aistats2024}
\usepackage[utf8]{inputenc} 
\usepackage[T1]{fontenc}    
\usepackage{hyperref}       
\usepackage{url}            
\usepackage{booktabs}       
\usepackage{amsmath}
\usepackage{amssymb}
\usepackage{mathtools}
\usepackage{amsthm}      
\usepackage{nicefrac}       
\usepackage{microtype}      
\usepackage{lipsum}		
\usepackage{graphicx}
\usepackage{subfigure}
\usepackage{bbm}
\usepackage{natbib}
\usepackage{algorithm}
\usepackage{algorithmic}

\usepackage{doi}
\setcitestyle{authoryear,round}
\newcommand*{\bv}[1]{\ensuremath \boldsymbol{#1}}

\newcommand*{\Xb}{\bv{X}}

\newcommand*{\xb}{\bv{x}}
\newcommand*{\yb}{\bv{y}}

\newcommand*{\thb}{\bv{\theta}}%
\newcommand*{\phb}{\bv{\phi}}%
\newcommand*{\psb}{\bv{\psi}}%
\begin{document}

%

%
\runningauthor{Ghosh, Birrell, De Angelis}

\twocolumn[

\aistatstitle{Sample-efficient neural likelihood-free Bayesian inference of implicit HMMs}

\aistatsauthor{ Sanmitra Ghosh$^{1,2}$ \\ \And Paul J. Birrell$^{3,2}$ \And Daniela De Angelis$^{2,3}$ \\ }

\aistatsaddress{  \texttt{\{sanmitra.ghosh,paul.birrell,daniela.deangelis\}@mrc-bsu.cam.ac.uk}\\$^{1}$PhysicsX, London, $^{2}$MRC Biostatistics Unit, University of Cambridge, $^{3}$UK Health Security Agency} ]

\begin{abstract}
  Likelihood-free inference methods based on neural conditional density estimation were shown to drastically reduce the simulation burden in comparison to classical methods such as ABC. When applied in the context of any latent variable model, such as a Hidden Markov model (HMM), these methods are designed to only estimate the parameters, rather than the joint distribution of the parameters and the hidden states. Naive application of these methods to a HMM, ignoring the inference of this joint posterior distribution, will thus produce an inaccurate estimate of the posterior predictive distribution, in turn hampering the assessment of goodness-of-fit. To rectify this problem, we propose a novel, sample-efficient likelihood-free method for estimating the high-dimensional hidden states of an implicit HMM. Our approach relies on learning directly the intractable posterior distribution of the hidden states, using an autoregressive-flow, by exploiting the Markov property. Upon evaluating our approach on some implicit HMMs, we found that the quality of the estimates retrieved using our method is comparable to what can be achieved using a much more computationally expensive SMC algorithm.
\end{abstract}
   
\section{INTRODUCTION}
We consider the task of carrying out Bayesian inference of an \textit{implicit} HMM, i.e. a HMM whose likelihood density function is analytically intractable. Such a model is only available as a simulator from which one can generate data that realistically, faithfully mimics the observed time course of some complex bio-physical system.   

Due to the analytical intractability of the likelihood density function standard Bayesian inference methods cannot be applied to such an implicit HMM. Inference of such a model is typically carried out using approximate Bayesian computation (ABC) \citep{sisson2018handbook}, which only requires forward simulations from the model, see for example \cite{martin2019auxiliary,picchini2014inference,Toni2009}. 

Recently, a new class of likelihood-free inference methods, see \cite{cranmer2020frontier} for a review, were developed that use a neural network based emulator of the posterior density, the likelihood density and the likelihood ratio. Such methods were empirically shown to be much more sample-efficient (require fewer model simulations) \citep{lueckmann2021benchmarking} than ABC. 
These methods were found to perform equally well across different models without problem specific tailoring of the neural network's architecture. Naturally, these methods appear as more preferable algorithmic choices for carrying out inference of an implicit HMM, in comparison to ABC. 

These \textit{neural likelihood-free} inference (NLFI) approaches, in the specific context of a latent variable problem such as a HMM, have so far been applied to carry out Bayesian inference partially by estimating only the marginal posterior of the parameters rather than the  joint posterior of the hidden states and parameters. This is since a naive implementation of a neural network based density (or density-ratio) estimator may perform unreliably (with drastically reduced sample-efficiency) in estimating the joint posterior of the parameters and the high-dimensional hidden states, potentially for a lack of inductive biases. Estimation of the hidden states may or may not be of interest within a particular application domain. However, without estimating the joint posterior of the parameters and the hidden states the goodness-of-fit cannot be correctly assessed. This is a severe limitation. Note that although ABC theoretically targets the joint distribution (see Appendix A.1 for details) it fails to estimate the hidden states adequately within a reasonable simulation budget. 

Note that the task of inferring the posterior of the hidden states of an implicit HMM is in itself extremely challenging. This problem can only be solved using the Bootstrap sequential Monte Carlo (SMC) algorithm \citep{gordon1995bayesian}. However, reliable performance of the Bootstrap SMC algorithm often requires a large number of model simulations, thus defeating the purpose of sample-efficient inference which motivates the use of NLFI. Thus, extending NLFI approaches to solve the joint estimation problem is highly non-trivial as this requires the development of a fundamentally new approach of estimating the hidden-states, that is much more sample-efficient in comparison to SMC. 

In this paper we present such a novel technique which is based on learning an approximation of the posterior distribution of the hidden states using neural density estimation. After learning this posterior approximation, neural density estimators can be used to draw the full path of the hidden states recursively. Following are our salient contributions:
\begin{itemize}
    \item We develop a sample-efficient method to obtain an approximation of the posterior distribution of the sample path of a HMM, without accessing the transition or the observation densities. 
    
    \item  Our approach, when combined with any off-the-shelf NLFI method, can be used, as a sample-efficient alternative to ABC, for carrying out full Bayesian inference of an implicit HMM.
    
\end{itemize}

\section{BACKGROUND}
We begin by first introducing the implicit HMM and then we will discuss the challenges of carrying out Bayesian inference. We can describe a HMM, for a latent Markov process $\Xb_t\in \mathbb{R}^K$ with a $K$-dimensional continuous state-space, as follows:
\begin{equation}\label{eq: HMM defn}
    \begin{aligned}
    \Xb_t \sim f(\Xb_t|\Xb_{t-1},\thb_f),\quad  \yb_t \sim g(\yb_t|\Xb_t,\thb_g)
    \end{aligned}
\end{equation}
where $\thb_f,\thb_g$ parameterise the transition $f(\Xb_t|\Xb_{t-1},\thb_f)$ and the observation $g(\yb_t|\Xb_t,\thb_g)$ densities respectively. We consider the parameter vector $\thb=(\thb_f,\thb_g,\Xb_0)$ to include the initial state $\Xb_0$. Given a set of noisy observations of $L$ out of the $K$ states $\yb \in \mathbb{R}^{M\times L}$ at $M$ experimental time points, of the latent process, our goal is to infer the joint posterior distribution $p(\thb,\xb|\yb)$, where $\xb= (\Xb_1,\ldots, \Xb_{M-1})$ is the unobserved sample path of the process -- the hidden states. The expression for the unnormalised posterior is given by
\begin{equation}\label{eq: True HMM posterior joint}
\small
\begin{aligned}
    p(\thb,\xb|\yb) \propto \Bigg(\prod_{t=0}^{M-1} g(\yb_t|\Xb_{t},\thb_g)\Bigg)
   & \Bigg(\prod_{t=1}^{M-1} f(\Xb_t|\Xb_{t-1},\thb_f)\Bigg)\\
   & \times p(\thb),
\end{aligned}
\end{equation}
where $p(\thb)$ is the prior distribution over the parameters and the initial values. Additionally, we are also interested in checking the goodness-of-fit, which, within the Bayesian context, is carried out by inspection of the posterior predictive distribution $p(\yb^r|\yb)$ of generating \textit{replicated data} \citep{gelman1996posterior} $\yb^r$. This distribution is given by
\begin{equation}\label{eq: ppc}
    p(\yb^r|\yb) = \int p(\yb^r|\xb,\thb)p(\xb,\thb|\yb)d\xb d\thb.
\end{equation} 

We assume that one can draw samples from $f(\cdot)$ and $g(\cdot)$, but cannot evaluate either or both of these densities. This assumption leads to an intractable likelihood density rendering the model implicit. This is the constrained setting for our work. Inference of $p(\thb,\xb|\yb)$, in our implicit modelling context, can be carried out using the ABC algorithm which replaces the evaluation of the right hand side of Eq \eqref{eq: True HMM posterior joint}, upto a normalising constant, by using a distance function between simulated and real data. Although ABC jointly samples (Appendix A.1) the hidden states and the parameters, drawing the high-dimensional hidden states just using \textit{rejection sampling} is highly inefficient. Although a more efficient variant of the basic ABC algorithm may employ a sophisticated technique to propose values of $\thb$, the states are still updated using the prior of the Markov process as the proposal, thus falling back to rejection sampling, resulting in an exorbitant computational expense. Due to this computational burden ABC algorithms are rarely practically useful for inference of an implicit HMM where $f(\cdot)$ and $g(\cdot)$ are computationally expensive simulators. 

Note that we can decompose the joint density using the product rule as follows:
\begin{equation}\label{eq:product}
    p(\xb,\thb|\yb) = p(\xb|\thb,\yb)p(\thb|\yb).
\end{equation}
With the above decomposition we can break down the task of inferring the joint distribution of $\xb,\thb$ into two sub-tasks of inferring separately the distributions $p(\thb|\yb)$ and $p(\xb|\thb,\yb)$. Samples of $\xb$ can then be drawn given samples of $\thb$. Note that the task of inferring $p(\thb|\yb)$ can be carried out, sample-efficiently, using any NLFI method. Inference of $p(\xb|\thb,\yb)$ can then be carried out using a Bootstrap SMC, or its ABC (and more inefficient) variant \citep{drovandi2016exact} when $g(\cdot)$ is unavailable. The posterior predictive distribution can then be evaluated as follows:
\begin{equation}\label{eq:smcppc}
    p(\yb^r|\yb) 
    \approx \int p(\yb^r|\xb,\thb)p_{smc}(\xb|\yb,\thb)p(\thb|\yb)d\xb d\thb,
\end{equation}
where $p_{smc}(\xb|\yb,\thb)$ is a Bootstrap SMC estimate of the hidden states. To evaluate the above integral numerically, one has to run a particle filter for each $\thb$ sample, which in turn will require as many simulations as the number of particles, resulting in a computation cost that in most cases will be higher than that of running NLFI for inferring $\thb$ alone. Clearly, this makes SMC unusable as long as we need to evaluate the posterior predictive distribution. The particle Markov chain Monte Carlo (MCMC) algorithm \citep{andrieu2010particle}, when $g(\cdot)$ is known, can produce samples from the true joint posterior distribution in Eq. \eqref{eq: True HMM posterior joint}. However, this algorithm also requires running SMC for each iteration of MCMC, making it computationally expensive to apply to complex models. Recently, neural network based methods have been proposed to infer a high-dimensional hidden states of a HMM \citep{schumacher2023neural,ryder2021neural}. However, it is unclear, given lack of comparison with exact algorithms, whether such methods can indeed recover the true posterior hidden states.


Note that standard sample-efficient alternatives to SMC such as EM family algorithms and more generally variational inference algorithms for state-space models are non-applicable due the implicitness of our model. Next, we briefly describe existing NLFI methods and highlight their limitations in estimating the joint posterior of $\xb,\thb$, before explaining the proposed method.


\subsection{Related work: Neural likelihood-free inference (NLFI)}\label{sec: Neural likelihood-free inference}

If instead of the joint $p(\thb,\xb|\yb)$ we only wish to estimate the marginal $p(\thb|\yb)$, then a number of strategies based on conditional density estimation can be employed. For example, we can simulate pairs of $\thb, \yb$ from their joint distribution and then subsequently create a training dataset, of $N$ samples $\{\thb^n,\yb^n\}_{n=1}^N$, which can be utilised to train a conditional density estimator, constructed using a flexible function approximator such as a neural network, that can approximate the marginal posterior \citep{papamakarios2016fast} $p(\thb|\yb)\approx q_{\psb}(\thb|\yb)$ or the likelihood \cite{papamakarios2019sequential} $p(\yb|\thb)\approx q_{\psb}(\yb|\thb)$. Here $\psb$ denotes the parameters of the function approximator used to build the density estimator. In the former case once we have trained an approximation to the posterior, using a density estimator, we can directly draw samples $\thb \sim q_{\psb}(\thb|\yb_o)$ by conditioning on a particular dataset $\yb_o$. In the latter case we can use the trained density estimator of the likelihood to approximate the posterior $p(\thb|\yb)\propto q_{\phb}(\yb_o|\thb)p(\thb)$ and then draw samples from it using MCMC. 

A neural network is used in this context either as a nonlinear transformation of the conditioning variables, within a mixture-of-Gaussian density as was proposed in \cite{bishop1994mixture}, or as a \textit{normalizing-flow} \citep{rezende2015variational,papamakarios2021normalizing} that builds a transport map \citep{parno2015transport} between a simple distribution (such as a standard Gaussian) and a complex one such as the likelihood/posterior density. Following the seminal work of \cite{tabak2013family} a large amount of research is undertaken to build such transport maps using samples from the respective measures. 

An alternative formulation of NLFI utilises the duality \citep{cranmer2015approximating} between the optimal decision function of a probabilistic classifier and the likelihood ratio, $r(\thb^a,\thb^b)=\frac{p(\yb|\thb^a)}{p(\yb|\thb^b)}$ evaluated using two samples $\thb^a$ and $\thb^b$, to approximate the likelihood-ratio through training a binary classifier using samples from $p(\yb,\thb)$. This likelihood ratio can then be used as proxy within a MCMC accept/reject step as follows:
\begin{equation}
\begin{aligned}
     \operatorname{min} \Bigg\{1,r(\thb^*,\thb)\frac{k_{\thb}(\thb|\thb^*)p(\thb^*)}{k_{\thb}(\thb^*|\thb)p(\thb)}\Bigg\},
    \end{aligned}
\end{equation}
where $k_{\thb}(\cdot)$ is a proposal density. Note that these NLFI methods carry out \textit{amortised inference} that is there is no need to re-learn the density/density-ratio estimator for every new instances of the observations. However, we like to point out that the MCMC algorithms, associated with likelihood or likelihood ratio estimation based approaches, has to be re-run for each new dataset, which can be more time consuming than training the associated neural networks.

To increase sample-efficiency of these methods one can use them in a sequential manner \citep{durkan2018sequential}. After an initial round of NLFI, we are left with samples of $\thb$ from its posterior distribution. We can subsequently use these samples to generate further simulated data concentrated around the given observations $\yb_o$. This constitute a new training set on which a second round of NLFI can be applied to further refine the approximations. This process can be repeated for a number of rounds. Note that when a sequential process is used in conjunction with a density estimator for the posterior then the parameter samples from the second round are no longer drawn from the prior. Thus, different adjustments had been proposed, leading to different algorithms \citep{greenberg2019automatic,lueckmann2017flexible}, to overcome this issue.

We like to further point out that NLFI is applied using some summary statistic $s(\yb)$ of the data $\yb$, a practise carried over from the usage of ABC methods. There have been recent work \citep{neurSum} on using a neural network to generate the summary statistics. 

Also see Appendix H for a brief review of other classical approaches for inference of implicit HMMs.

\section{LIMITATIONS OF NLFI METHODS}\label{limit nlfi}
\paragraph{Can we ignore the joint distribution?}

The various NLFI techniques discussed so far are designed to solve the marginal problem of estimating $p(\thb|\yb)$. The necessity of the estimation of the hidden states $\xb$ is problem specific. In some applications estimation of the hidden states is of paramount importance whereas in others one may wish to ignore the hidden states. Irrespective of whether the interest is in estimating $p(\thb,\xb|\yb)$ or $p(\thb|\yb)$, in the process of modelling a physical phenomenon it is necessary to assess the goodness-of-fit.
But when instead of the joint distribution we only have access to the marginal distribution (that is access to only samples of $\thb$, the output of any NLFI method) then the posterior predictive distribution, instead of Eq.~\eqref{eq: ppc},  can only be obtained as follows
\begin{equation}\label{eq: ppcwrng}
    p(\yb^r|\yb) \approx \hat{p}(\yb^r|\yb) = \int p(\yb^r|\xb,\thb)p(\xb|\thb)p(\thb|\yb)d\xb d\thb,
\end{equation}
where the joint posterior of $\xb,\thb$ is approximated as $p(\xb,\thb|\yb)\approx p(\xb|\thb)p(\thb|\yb)$, which is akin to drawing $\xb$ from the prior $p(\xb|\thb)=\prod_{t=1}^{M-1} f(\Xb_{t}|\Xb_{t-1},\thb)$, of the latent Markov process. As a result the credible intervals of $\hat{p}(\yb^r|\yb)$ would be erroneously inflated, since in this case the latent sample path $\xb$ is not correctly constrained by the data, leading to an incorrect assessment of the goodness-of-fit. This is a severe problem that needs to be addressed even in the case where we wish to estimate just the marginal $p(\thb|\yb)$.

\paragraph{limitations of NFLI for inferring the joint:}

let us now consider the task of estimating the joint posterior $p(\thb, \xb|\yb)$ using a NLFI method. If we want to approximate the posterior then we have to extend any chosen density estimator to target a high-dimensional vector $(\thb, \operatorname{vec}(\xb))$, where $\operatorname{vec}:\mathbb{R}^{K\times M}\rightarrow \mathbb{R}^{K M}$, which would invariably require a larger training set, and thus more simulations, in comparison to estimating only $\thb$ (see Appendix F for an example). Alternatively, if we choose to approximate the likelihood density, then note that the accept/reject step of a MCMC scheme, targeting $p(\xb,\thb|\yb)$, will be of the following form:
\begin{equation}
\small
    \operatorname{min} \Bigg\{1,\frac{q_{\psb}(\yb_o|\xb^*,\thb^*)p(\xb^*|\thb^*)p(\thb^*) k_{\xb}(\xb|\xb^*)k_{\thb}(\thb|\thb^*)}{q_{\psb}(\yb_o|\xb^,\thb)p(\xb|\thb)p(\thb) k_{\xb}(\xb^*|\xb)k_{\thb}(\thb^*|\thb)}\Bigg\},
\end{equation}
where $k_{\xb}(\cdot)$, $k_{\thb} (\cdot)$ are the proposal densities. Due to the intractability of $p(\xb,\thb)=p(\xb|\thb)p(\thb)$ our only option as a proposal, $k_{\xb}(\cdot)$, is the prior (so that the proposal and prior of $\xb$ cancel out in the above ratio) that is the transition density in equation \ref{eq: HMM defn}. This will jeopardise the mixing of the MCMC sampler which in turn would require excessive simulation from the model. We would face the same limitation if we had chosen to emulate the likelihood ratio. 

\section{METHODS}

\subsection{Inferring hidden states}
\label{sec:Learning an incremental posterior}


We can decompose the posterior of $\xb$, by applying the product rule and then utilising the Markov property (see proof in Appendix A.2), as follows:
\begin{equation}\label{eq:main approximation1}
\begin{aligned}
p(\xb|\thb,\yb) &= p(\Xb_{M-1}|\Xb_{M-2},\thb,\yb_{M-1})\\
&\times \prod_{t=1}^{M-2} p(\Xb_{t}|\Xb_{t+1}, \Xb_{t-1}, \yb_t, \thb).    
\end{aligned}
\end{equation}
Note that the above decomposition produces homogeneous factors $p(\Xb_{t}|\Xb_{t+1}, \Xb_{t-1}, \yb_t, \thb)$. By dropping the dependency of the future sample point $\Xb_{t+1}$ and thus approximating each factor,  
\begin{equation}
    p(\Xb_{t}|\Xb_{t+1}, \Xb_{t-1}, \yb_t, \thb) \approx p(\Xb_{t}|\Xb_{t-1}, \yb_t, \thb),
\end{equation}
we can approximately decompose the posterior as follows:
\begin{equation}\label{eq:main approximation}
    p(\xb|\thb,\yb) \approx \prod_{t=1}^{M-1} p(\Xb_{t}|\Xb_{t-1}, \yb_t, \thb).
\end{equation}
 Although we are losing information, this approximation will still be a reasonable one as long as the information in $\Xb_{t+1}$ is largely contained in the pair $(\Xb_{t-1}, \yb_{t})$. 
 Importantly, this approximation lets us easily draw the hidden states using ancestral sampling from the approximate factor $p(\Xb_{t}|\Xb_{t-1}, \yb_t, \thb)$. Additionally, we can improve this approximation by employing importance sampling. That is we can obtain a sample from the correct factor $p(\Xb_{t}|\Xb_{t+1}, \Xb_{t-1}, \yb_t, \thb)$ by drawing a weighted sample from the approximate one $p(\Xb_{t}|\Xb_{t-1}, \yb_t, \thb)$, with weights given by
 \begin{equation}
     w_t ( \Xb_{t}) = \frac{p(\Xb_{t}|\Xb_{t+1}, \Xb_{t-1}, \yb_t, \thb)}{p(\Xb_{t}|\Xb_{t-1}, \yb_t, \thb)}.
 \end{equation}

 Having introduced a technique for drawing the hidden states we must point out that except for linear-Gaussian models, these factors are never available in closed form. Thus, the decomposition in Eq.~\eqref{eq:main approximation1}, in our knowledge, has never been explored in the context of classical filtering/smoothing methodologies. 
 
 Since these factors are homogeneous thus we can now emulate the approximate and true: 
 \begin{equation}\label{eq:ide estimators}
 \begin{aligned}
 \small 
     p(\Xb_t| \Xb_{t-1}, \yb_{t}, \thb) &\approx q_{\phb_1}(\Xb_t| \Xb_{t-1}, \yb_{t}, \thb)\\
     p(\Xb_t|\Xb_{t+1}, \Xb_{t-1}, \yb_{t}, \thb)&\approx q_{\phb_2}(\Xb_t|\Xb_{t+1}, \Xb_{t-1}, \yb_{t}, \thb),
 \end{aligned}
 \end{equation}
 factors between any two consecutive time points $t,t-1$, using neural density estimators $q_{\phb_1}(\cdot),q_{\phb_2}(\cdot)$, where $\phb_1,\phb_2$ are the parameters of the respective neural networks, parameterising the density estimators in turn. With access to these neural density estimates of the correct and approximate factors, an approximation to the entire sample path can be generated again using importance sampling where weighted samples $\Xb_t$ can be recursively drawn at every time point, with weights given by
 \begin{equation}\label{eq: weights}
     w_t ( \Xb_{t})= \frac{q_{\phb_2}(\Xb_{t}|\Xb_{t+1}, \Xb_{t-1}, \yb_t, \thb)}{q_{\phb_1}(\Xb_{t}|\Xb_{t-1}, \yb_t, \thb)}.
 \end{equation}
 In practise the above importance sampling can be carried out in two steps. First, we draw a cloud of $P$ particles, at each time point, from the importance factor: $\hat{\Xb}^p_t \sim q_{\phb_1}(\cdot| \hat{\Xb}^p_{t-1}, \yb_{t}, \thb)$, and assign weight $w^p_t:=w(\hat{\Xb}^p_t)$ to each particle $p=1,\ldots,P$. We then resample a single $\Xb_t$, at each $t$ from the particle cloud to construct the desired sample path $\xb$.

Since we are using a density estimator, rather than the HMM itself, as above, we can use a large number of importance samples (we chose $P=10^4$, throughout), unlike traditional SMC algorithms, without bothering about computational cost.
 
\begin{figure*}
\includegraphics[width=1\textwidth,height=1\textheight,keepaspectratio=true]{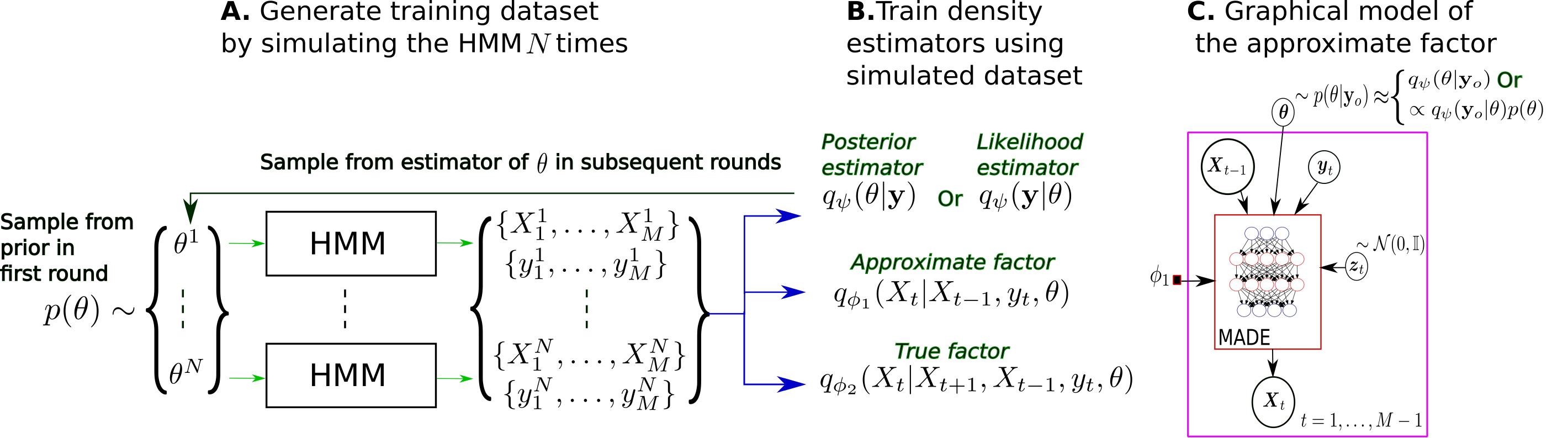}
\caption{The process of using neural density estimators to approximate the joint posterior distribution $p(\thb, \xb|\yb)$ of a HMM. First, the HMM (simulator) is used to generate a training dataset $\{\thb^n, \xb^n, \yb^n\}_{n=1}^N$ (\textbf{A}), which is then used to train three neural density estimators (\textbf{B}). Training of the estimators of $\thb$ happens sequentially through multiple rounds, generating more simulated data in the process. Once trained, given an observed time series $\yb_{o}$, for each posterior sample of $\thb$ drawn using its estimator, the approximate factor recursively generates (\textbf{C}) \textit{importance samples} of the latent path. The hidden states are resampled from these importance samples using weights that are the ratio of the true and approximate factors (see Algorithm \ref{alg:IDE pred} for the pseudocode). }
\label{training}
\end{figure*}

 We denote the above strategy (see Algorithm \ref{alg:IDE pred} for the pseudocode) of drawing the hidden states $\xb$ using neural density estimates of the true and approximate factors (both being essentially an incremental posterior), collectively as an \textit{incremental density estimator} (IDE). Using the IDE we can approximate the posterior predictive in \eqref{eq: ppc} now as follows:
\begin{equation}\label{eq:ppcide}
\begin{aligned}
    p(\yb^r|\yb) &\approx \int p(\yb^r|\xb,\thb)
q_{\phb_1}(\Xb_{M-1}|\Xb_{M-2},\thb,\yb_{M-1})\\
& \prod_{t=1}^{M-2}  w_t ( \Xb_{t})q_{\phb_1}(\Xb_{t}|\Xb_{t-1}, \yb_t, \thb)
 p(\thb|\yb)d\xb d\thb.
\end{aligned}
\end{equation}
\begin{algorithm}[htb!]
   \caption{Hidden states prediction using IDE}
   \label{alg:IDE pred}
\begin{algorithmic}
   \STATE {\bfseries Input:} Posterior parameter samples $\{\thb^{l}\}_{l=1}^L$, observed time series $\yb_{o}=(\yb_{o_1}, \ldots,\yb_{o_M})$ of length $M$, number of particles $P$, neural density estimators $q_{\phb_1}(\cdot),q_{\phb_2}(\cdot)$ introduced in Eq \eqref{eq:ide estimators}.\\
 \STATE 1. Generate importance samples
   \FOR{$l=1$ {\bfseries to} $L$}   
   \FOR{$t=1$ {\bfseries to} $M-1$}
   \FOR{$p=1$ {\bfseries to} $P$}
   \STATE Draw importance samples of the hidden states
   $\hat{\Xb}^{l,p}_t \sim q_{\phb_1}(\cdot|\hat{\Xb}^{l,p}_{t-1}, \yb_{o_t}, \thb^l)$.
   \STATE Obtain importance weights\\
   $w^{l,p}_t(\hat{\Xb}^{l}_t) = 
   \frac{q_{\phb_2}(\hat{\Xb}^{l,p}_{t}|\hat{\Xb}^{l,p}_{t+1}, \hat{\Xb}^{l,p}_{t-1},\yb_{o_t}, \thb^l)}{q_{\phb_1}(\hat{\Xb}^{l,p}_{t}|\hat{\Xb}^{l,p}_{t-1},\yb_{o_t}, \thb^l)}
   $.
   \ENDFOR
   \STATE Normalise the weights, set $w^{l,p}_t = \frac{w^{l,p}_t}{\sum_{p=1}^P w^{l,p}_t}$
   \ENDFOR
    \ENDFOR
    \STATE 2. Generate weighted samples
    \FOR{$l=1$ {\bfseries to} $L$}
    \FOR{$t=1$ {\bfseries to} $M-1$}   
   \STATE Resample an index $r$ from the set $\{1,\ldots, P\}$, with respective weights $\{w^{l,1}_t,\ldots, w^{l,P}_t \}$.
   \STATE Set $\Xb^{l}_{t}=\hat{\Xb}^{l,r}_{t}$.      
   \ENDFOR
    \ENDFOR
   \STATE {\bfseries Output:} Hidden states $\Xb\in \mathbb{R}^{M \times L}$.
\end{algorithmic}
\end{algorithm}
In summary our strategy, see Figure~\ref{training} for an overview, for drawing samples $\xb,\thb$ from an approximation of their joint posterior is as follows. We first infer the parameter marginal $p(\thb|\yb)$ using any chosen off-the-shelf NLFI method (see section \ref{sec: Neural likelihood-free inference}) and draw samples of $\thb$. In parallel we train an IDE using a subset of simulations used in inferring $p(\thb|\yb)$. Then for each sample $\thb$, and an observed dataset $\yb_{o}$, we can use the trained IDE recursively to obtain the sample path $\xb$ conditioned on the sample $\thb$ and the dataset $\yb_{o}$.  

\paragraph{Limitations:} There are two fundamental assumptions behind our approach. Firstly, we assume that samples of $\thb$ are drawn from a good approximation to the true unknown posterior. However, this may not be true when inference of $\thb$ is done on a very limited simulation budget. Secondly, we assume that there is no model miss-specification and the actual observations come from the joint distribution $p(\xb,\thb,\yb)$ used for learning the IDE. This can be a strong assumption when modelling a new phenomenon. 

\subsection{Training the incremental density estimator}\label{sec: IDE training description}
For simplicity we will provide details about the neural density estimation for the approximate factor $q_{\phb_1}(\Xb_t| \Xb_{t-1}, \yb_{t}, \thb)$, from which we can draw the sample path recursively. The corresponding neural density estimator for the true factor can be constructed and trained analogously.

We chose a \textit{masked autoregressive flow} (MAF) as the incremental neural density estimator. MAF is built upon the idea of chaining together a series of autoregressive functions, and can be interpreted as a normalizing-flow \citep{papamakarios2017masked}. That is 
we can represent  $q_{\phb_1}(\Xb_t|\Xb_{t-1}, \yb_t, \thb)$ as a transformation of a standard Gaussian $\mathcal{N}(\bv{0},\mathbb{I})$ (or another simple distribution) through a series of $J$ autoregressive functions $h^1_{\phb_1},\ldots,h^J_{\phb_1}$, parameterised by $\phb_1$, each of which is dependent on the triplet $(\Xb_{t-1}, \yb_{t}, \thb)$:
\begin{equation}
    \Xb_{t} = \bv{z}_J, \quad \operatorname{where} \quad \begin{array}{l}
    \bv{z}_0 = \mathcal{N}(\bv{0},\mathbb{I})\\
     \bv{z}_j = h^j_{\phb_1}(\bv{z}_{j-1},\Xb_{t-1}, \yb_{t}, \thb)
  \end{array}.
\end{equation}
Each $h_j$ is a bijection with a lower-triangular Jacobian matrix, implemented by a \textit{Masked Autoencoder for Distribution Estimation} (MADE) \citep{germain2015made}, and is conditioned on $(\Xb_{t-1}, \yb_t, \thb)$. Using the formula for change of variable the density is given by
\begin{equation}
   q_{\phb_1}(\Xb_t|\Xb_{t-1}, \yb_t, \thb) = \mathcal{N}(\bv{0},\mathbb{I}) \prod_{j=1}^J \Biggm\lvert \det \left(\frac{\partial h^j_{\phb_1}}{\partial \bv{z}_{j-1}}\right)\Biggm\lvert^{-1}.
\end{equation}

We can learn the parameters $\phb_1$ by maximising the likelihood. To do this we create a training dataset consisting of $N$ examples. We first sample $N$ values of the parameter $\{\thb^n\}_{n=1}^N$ from its prior and for each $\thb^n$ we simulate the sample path of the states and the observations using \eqref{eq: HMM defn}. Each training example is then created by collecting the random variables: $(\Xb^n_i, \yb^n_j, \thb^n)$, and $\Xb^n_j$, as the input-target pair, where $(i,j)= (0,1), (2,3), \ldots, (M-2,M-1)$. Clearly, even with a small number of model simulations (a few thousands) we can create a large training dataset to learn an expressive neural density estimator. 

In Figure~\ref{training} we outline the process of creating this training dataset. Given these training examples $\phb$ can be learnt, using gradient ascent, through maximising the total likelihood:
\begin{equation}
    \mathcal{L}(\phb_1)=\sum_{n=1}^N\sum_{i=0,j=1}^{M-2,M-1} \log q_{\phb_1}(\Xb^n_j|\Xb^n_i, \yb^n_j, \thb^n),
\end{equation}
which is equivalent to minimising the forward Kullback–Leibler divergence $\operatorname{KL}\left(p(\Xb_t|\Xb_{t-1}, \yb_t, \thb)|| q_{\phb_1}(\Xb_t|\Xb_{t-1}, \yb_t, \thb)\right)$ \citep{papamakarios2019sequential} between the approximate factor and its neural density estimate. 

Pseudocode of sampling and training of the IDE is provided in Appendix A.3.

\section{EVALUATIONS}
\label{Evaluations}
We evaluated the proposed approach in two stages. First, we used a nonlinear Gaussian state-space model which has a tractable approximate factor $p(\Xb_t|\Xb_{t-1},\yb_t, \thb)$. This tractability lets us compare the IDE with a conditionally optimal SMC algorithm (more accurate than bootstrap SMC). In the next stage, we used two implicit biological HMMs models to evaluate the IDE's usefulness in accurately estimating the hidden states and the posterior predictive distribution.

\subsection{State-space model with a tractable approximate factor}
We considered a state-space model, that has a tractable approximate factor, to evaluate the quality of approximation of the hidden states $\xb\sim p(\xb|\yb,\thb)$ in a classical \textit{filtering context} \citep{sarkka2013bayesian}. Specifically, we used the following model:
\begin{equation}
    \begin{aligned}
        \Xb_t &\sim \mathcal{N}(\bv{A} \gamma(\Xb_{t-1} ), \sigma^2_x \mathbb{I}) \quad t\geq 1\\
        \yb_t &\sim \mathcal{N}(\bv{B} \Xb_t, \sigma^2_y \mathbb{I}),
    \end{aligned}
\end{equation}
where $\gamma(\Xb )=\sin (\exp(\Xb_{t-1}))$, applied elementwise, $\bv{A}=\mathbb{I}_{K\times K}$, $B=2\bv{A}$ and $\Xb_0=\bv{0}$. 

We considered a moderately high-dimensional state-space, $K=L=10$, with a long enough time series, $M=500$, to challenge SMC algorithms. Naturally, a model setup that is challenging for a SMC algorithm will suffice as a good test-bed for the IDE. We considered the parameters $\sigma_x,\sigma_y$ to be known and fixed, thus conditioning the IDE only on $(\Xb_{t-1},\yb_t)$.To estimate the sample path $\xb$ we applied the IDE (see Appendix B.1 for details of the neural networks used), the bootstrap SMC, and the conditionally optimal SMC \citep{doucet2000sequential}, also known as the guided SMC, that uses $p(\Xb_t|\Xb_{t-1},\yb_t)$, a Gaussian distribution, see Appendix B.1, as the importance proposal. Note that the guided SMC is not applicable in case of an implicit model. However, due to its superior performance, when we have a tractable $p(\Xb_t|\Xb_{t-1},\yb_t)$, we have used this algorithm as the \textit{gold standard} for this model.

The goal of this experiment was to find out how the quality of estimation, by methods that require simulation, vary with the number of simulations. Essentially comparing the sample-efficiency. For the IDE this is determined by the training set size, and for SMC the number of particles. We chose the (i) \textit{mean squared error} (MSE) between the true hidden states and it's posterior mean, to quantify the bias, and (ii) the $90 \%$ \textit{empirical coverage} (EC), of the ground truth, to quantify the quality of uncertainty estimation. 
\begin{figure}[!ht]
\includegraphics[width=0.489\textwidth]{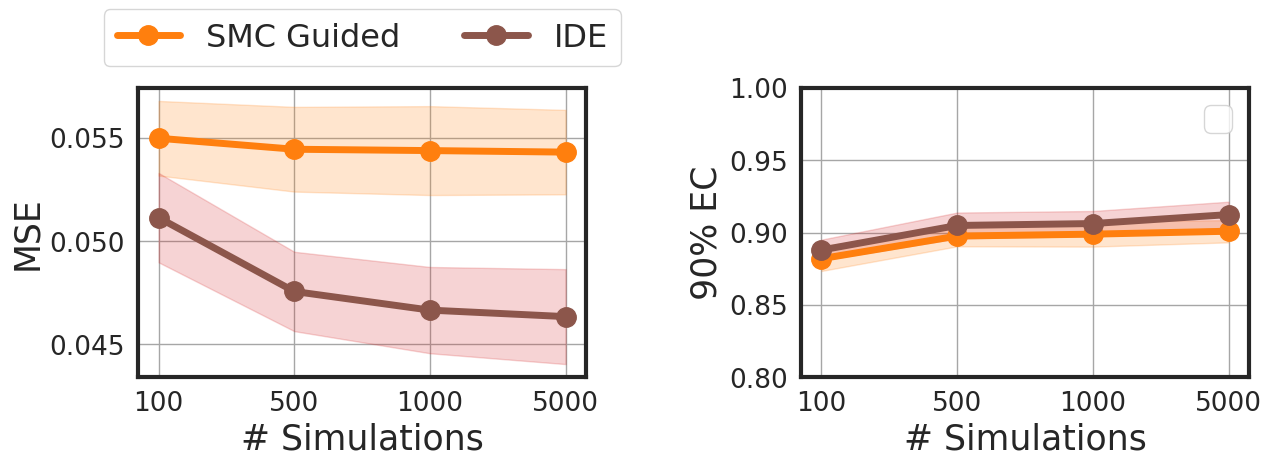}
\caption{Estimation of the \textbf{hidden states} of a \textbf{nonlinear state-space} model. The quality of approximations was quantified using the MSE and $90 \%$ EC, summarised using the mean (solid line) and $95 \%$ confidence intervals (shaded area), across $10$ datasets.}
\label{state-space}
\end{figure}
\begin{table*}
\caption{Estimates of the \textbf{hidden states} of the \textbf{Lotka-Volterra} and \textbf{Prokaryotic autoregulator} models. Except ABC-SMC, we denoted all other methods using the following convention: method to estimate $\xb$ (NLFI algorithm for estimating $\thb$). Metrics were summarised by mean $\pm$ standard deviation across $10$ simulated datasets. The baseline is SMC.}
\label{Esimates summaries hidden}
       
\resizebox{\textwidth}{!}{%
\begin{tabular}{lccc}
\toprule
\multicolumn{4}{c}{The \textbf{Lotka-Volterra} model} \\
\midrule
Methods & MSE   & $90 \%$ EC & CV \\
\midrule
ABC-SMC  & $3654.41\pm 2239.14$ & $0.99\pm 0.0091$ & $0.79\pm 0.16$  \\
PrDyn (SRE)  & $3585.03\pm 1938.88$ & $0.98\pm 0.02$ & $0.75\pm 0.16$  \\
PrDyn (SNLE)  & $4165.14\pm 1869.35$ & $0.97\pm 0.02$ & $0.81\pm 0.16$  \\
SMC (SRE)  & $57.19\pm 7.97$ & $0.93\pm 0.03$ & $0.07\pm 0.01$  \\
SMC (SNLE)  & $56.86\pm 7.79$ & $0.92\pm 0.03$ & $0.07\pm 0.01$  \\
IDE (SRE)  & $58.13\pm 6.11$ & $0.95\pm 0.01$ & $0.08\pm 0.01$  \\
IDE (SNLE)  & $57.85\pm 6.89$ & $0.95\pm 0.01$ & $0.08\pm 0.01$  \\
\bottomrule
\end{tabular}
\begin{tabular}{lccc}
\toprule
\multicolumn{4}{c}{The \textbf{Prokaryotic autoregulator} model} \\
\midrule
Methods & MSE   & $90 \%$ EC & CV \\
\midrule
ABC-SMC  & $17.28\pm 4.76$ & $0.99\pm 0.01$ & $1.03\pm 0.11$  \\
PrDyn (SRE)  & $19.71\pm 6.68$ & $0.99\pm 0.01$ & $0.80\pm 0.02$  \\
PrDyn (SNLE)  & $20.14\pm 7.51$ & $0.99\pm 0.01$ & $0.72\pm 0.04$  \\
SMC (SRE)  & $3.05\pm 0.61$ & $0.93\pm 0.03$ & $0.73\pm 0.02$  \\
SMC (SNLE)  & $2.98\pm 0.59$ & $0.94\pm 0.02$ & $0.65\pm 0.04$  \\
IDE (SRE)  & $2.83\pm 0.54$ & $0.97\pm 0.01$ & $0.72\pm 0.02$  \\
IDE (SNLE)  & $2.78\pm 0.46$ & $0.98\pm 0.01$ & $0.66\pm 0.04$  \\
\bottomrule
\end{tabular}
}
\end{table*}

In Figure~\ref{state-space} we plot these metrics, summarised across $10$ simulated datasets. The Bootstrap SMC performed poorly and its particle system completely degenerated within a few time steps resulting in a comparatively high MSE ($\geq 0.4$) for all the datasets. Thus, we avoid plotting the Bootstrap SMC results in Figure~\ref{state-space}. We found similar degeneration for other experimental settings (see Appendix B.2). The IDE outperformed the guided SMC algorithm, in terms of both the metrics. It produced a noticeably smaller MSE than what was produced by the guided SMC using the largest particle population, $5000$. Thus, it is clearly evident that although the guided SMC's importance proposal is better than a vanilla Bootstrap SMC, a large number of particles (much larger than the maximum, $5000$, used in this experiment) is required to make its performance comparable to the IDE. Especially, for a problem with a large value of $K \times M$. This experiment also shows that for complex high-dimensional implicit models, where guided SMC is inapplicable, IDE, using significantly fewer simulations, may perform at par or better than the bootstrap SMC.

\subsection{Implicit biological HMMs}
To evaluate the usefulness of the IDE for accurate estimation of the posterior predictive distribution $p(\yb^r|\yb)$, we considered two biological HMMs: (i) a stochastic Lotka-Volterra (LV) \citep{wilkinson2018stochastic}, and (ii) a prokaryotic autoregulator (PKY) \citep{golightly2011bayesian} model. Further details of the dynamics, data generation and priors can be found in Appendix C. The hidden states for these models evolve as a pure Markov jump process (MJP) and thus the density $f(\cdot)$ is unavailable. Throughout we used simulated data so that we are cognisant of the ground truth. We used a tractable observation density $g(\cdot)$ to facilitate the Bootstrap  SMC algorithm, run with $100$ particles following \cite{golightly2011bayesian}. SMC estimates were considered as the baseline. 

We chose the following competing approaches. First, the ABC-SMC algorithm \citep{Toni2009}, which produces samples from the joint distribution $p(\xb,\thb|\yb)$, that can be used to evaluate the posterior predictive in \eqref{eq: ppc}. Except ABC-SMC all other approaches rely on the availability of parameters samples from the marginal $p(\thb|\yb)$, which can be obtained using any off-the-shelf NLFI method such as the ones discussed in section \ref{sec: Neural likelihood-free inference}. Once samples of $\thb$ become available then samples of $\xb$ can be drawn from its posterior using the IDE, the SMC (since $g(\cdot)$ is available), or simply from the prior transition $p(\xb|\thb)$ (which we denote as PrDyn). Samples from $p(\yb^r|\yb)$ can then be drawn using \eqref{eq:ppcide} for IDE, \eqref{eq:smcppc} for SMC and \eqref{eq: ppcwrng} for PrDyn. 

To estimate $p(\thb|\yb)$, required for IDE, SMC and PrDyn approaches, we used two sequential NLFI methods. One based on learning the likelihood density (SNLE) and the other one based on learning the likelihood-ratio (SRE). It was recently shown in \citep{durkan2020contrastive} that SRE \citep{hermans2020likelihood} is equivalent to a certain form of sequential learning of the posterior density (SNPE-C) \citep{greenberg2019automatic} and both can be unified under a common framework on contrastive learning \citep{gutmann2010noise}. Thus, by using SNLE and SRE we can cover the general ambit of sequential NLFI approaches. Further details of the neural network architecture, optimisation and other relevant details for IDE/SNLE/SRE are given in Appendix D.
\begin{table*}
\caption{Estimates of the \textbf{posterior predictive distribution} of the \textbf{Lotka-Volterra} and \textbf{Prokaryotic autoregulator} models, summarised across $10$ simulated datasets. The baseline is SMC.}
\label{Esimates summaries ppc}
       
\resizebox{\textwidth}{!}{%
\begin{tabular}{lccc}
\toprule
\multicolumn{4}{c}{The \textbf{Lotka-Volterra} model} \\
\midrule
Methods & MSE   & $90 \%$ EC & CV \\
\midrule
ABC-SMC             & $3791.01\pm 2239.27$      & $0.98\pm 0.01$ & $0.80\pm 0.16$  \\
PrDyn (SRE)         & $3671.44\pm 1876.06$      & $0.97\pm 0.02$ & $0.76\pm 0.16$  \\
PrDyn (SNLE)        & $4260.82\pm 1827.63$      & $0.96\pm 0.02$ & $0.82\pm 0.16$  \\
SMC (SRE)           & $55.50\pm 13.08$          & $0.98\pm 0.01$ & $0.12\pm 0.02$  \\
SMC (SNLE)          & $54.61\pm 12.43$          & $0.98\pm 0.01$ & $0.12\pm 0.02$  \\
IDE (SRE)  & $100.69\pm 28.02$         & $0.96\pm 0.03$ & $0.12\pm 0.02$  \\
IDE (SNLE) & $99.21\pm 27.39$          & $0.96\pm 0.03$ & $0.12\pm 0.02$  \\
\bottomrule
\end{tabular}
\begin{tabular}{lccc}
\toprule
\multicolumn{4}{c}{The \textbf{Prokaryotic autoregulator} model} \\
\midrule
Methods & MSE   & $90 \%$ EC & CV \\
\midrule
ABC-SMC             & $22.23\pm 5.45$      & $0.98\pm 0.02$ & $0.34\pm 0.07$  \\
PrDyn (SRE)         & $24.87\pm 7.46$      & $0.98\pm 0.01$ & $0.33\pm 0.03$  \\
PrDyn (SNLE)        & $25.20\pm 8.48$      & $0.98\pm 0.02$ & $0.32\pm 0.03$  \\
SMC (SRE)           & $2.15\pm 0.46$       & $0.99\pm 0.01$ & $0.11\pm 0.01$  \\
SMC (SNLE)          & $1.76\pm 0.54$       & $0.99\pm 0.01$ & $0.11\pm 0.01$  \\
IDE (SRE)  & $2.36\pm 0.43$       & $0.99\pm 0.01$ & $0.12\pm 0.02$  \\
IDE (SNLE) & $2.05\pm 0.51$       & $0.99\pm 0.01$ & $0.12\pm 0.01$  \\
\bottomrule
\end{tabular}
}
\end{table*}
We used a fixed budget of simulations respectively for ABC-SMC (which jointly estimates $\thb,\xb$) and SNLE/SRE (used for estimating $\thb$). This was done to rule out major differences in the estimates of $\thb$ among ABC-SMC and NLFI methods so that the differences in estimates of $\xb$, and subsequently $\yb^r$, cannot be attributed to differences in parameter estimates. These simulation budgets were informed by previous studies such as \cite{lueckmann2021benchmarking} that compared the sample-efficiency between ABC-SMC and NLFI methods for estimating $p(\thb|\yb)$. Thus, for both models, while using SNLE/SNRE, we used $30$ rounds and the posterior samples from the final round were collected. For both models we generated $5000$ training examples in the first round and in the subsequent rounds we generated $1000$ examples. The simulations generated in the first round were used to learn the parameters $\phb$ of the IDE. We limited the ABC-SMC to use no more than $10^7$ simulations from the model. ABC-SMC is far less sample-efficient in comparison to SNLE/SRE \citep{lueckmann2021benchmarking}. Hence, we used considerably more simulations, in case of ABC-SMC, to ensure that the parameter estimates are as close as possible to SNLE/SRE. Further details of ABC-SMC implementation are given in Appendix D. 

For inferring the parameters using ABC, SNLE/SRE we used summary statistics chosen to preserve the dynamical properties (e.g limit cycles). Note that IDE and SMC require full data, and PrDyn does not use data. Additionally, these methods use the same $\thb$ values. Thus, these methods' relative performances are not influenced by the choice and use of summary statistics.  

As before we considered the MSE (between the ground truth and the posterior mean) and $90 \%$ EC as metrics to quantify the quality of estimates of the replicated data $\yb^r$ and the hidden states $\xb$. However, methods such as the PrDyn are bound to overestimate the uncertainty (see section \ref{limit nlfi}). Thus, we have also quantified the dispersion of $p(\yb^r|\yb)$ and $p(\xb|\thb,\yb)$ using \textit{coefficient of variation} (CV): the ratio of the posterior standard deviation and posterior mean, averaged across the time points. 

The NLFI approaches were implemented using the \texttt{sbi}\footnote{https://www.mackelab.org/sbi/} package \citep{tejero-cantero2020sbi}. We implemented the stochastic simulation algorithm \citep{gillespie1977exact} in \texttt{C++} to simulate the LV and PKY models. All the experiments were carried out on a high-performance computing
cluster. Our code is available at \texttt{https://github.com/sg5g10/HMM}.

\textbf{Additional experiments:}
In Appendix E we have also carried out evaluations on the PKY model without using summary statistics, using the SRE (which can learn summaries on the fly) and ABC-SMC, where we see no major differences in performance.
Additionally, we have also run an experiment with the LV model to show the perils of trying to  estimate $\xb,\thb$ jointly using a neural density estimator. See Appendix F for details.

\textbf{Results:}
 We quantified the quality of estimation of the posterior distribution of the hidden states in Table~\ref{Esimates summaries hidden} and subsequently the posterior predictive distribution in Table~\ref{Esimates summaries ppc}. In both these Tables, except ABC-SMC, we denoted all other methods using the following convention: \textbf{method to estimate $\xb$} (\textbf{NLFI algorithm for estimating $\thb$}). We summarised the chosen metrics across $10$ simulated datasets. Clearly, using the IDE we were able to produce an estimate of the hidden states and subsequently the posterior predictive that is closer or better (hidden states of PKY model) to what can be achieved using SMC (which is highly sample-inefficient). We noticed that all the methods were producing higher values of the empirical coverage. For methods such as PrDyn such higher values do not indicate good uncertainty quantification. Rather such high coverage, for these methods, indicate credible intervals that are wide enough to always contain the ground truth. This was verified upon inspecting the CV metric which indicated significantly higher dispersion, for PrDyn and ABC-SMC, methods which draws $\xb$ using its prior ($f(\cdot)$ ), indicating overestimation of the uncertainty. Notice the overestimation of uncertainty in plots of the posterior sample paths (Appendix G.1). PrDyn and ABC-SMC, by relying on the prior of the Markov process for proposing the hidden states, end-up producing a highly biased and under-confident estimate of the posterior distribution of the hidden states and subsequently the posterior predictive distribution.
\begin{figure}[!ht]

  \centering
  
  \begin{subfigure}[]{
    \includegraphics[width=.21\textwidth]{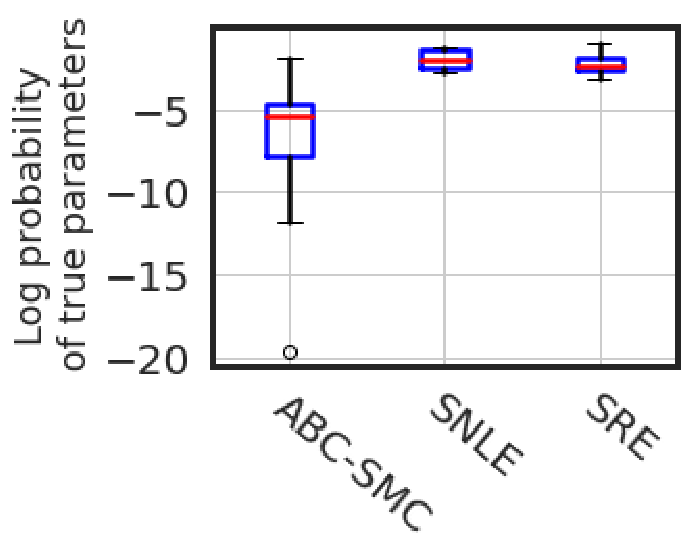}
    \label{Figure:lvlogprob}
  }  
  \end{subfigure}
	\begin{subfigure}[]{
    
    \includegraphics[width=.22\textwidth]{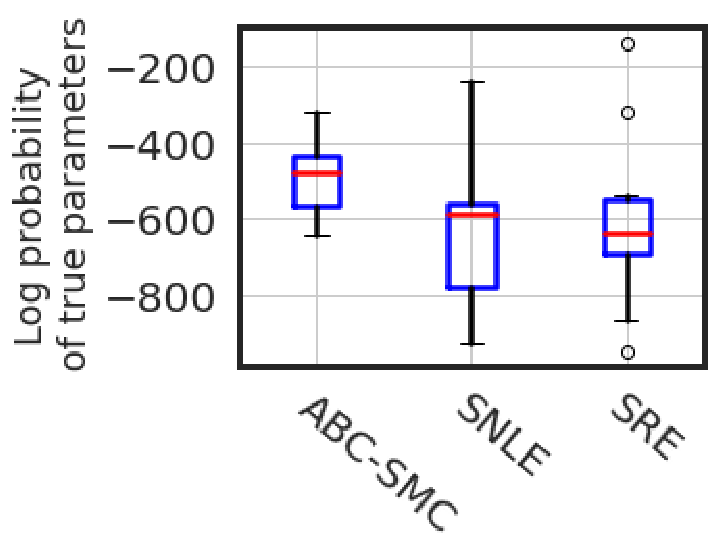}
    \label{Figure:pkylogprob}
}\end{subfigure}

\caption{Accuracy of parameter estimates for the \textbf{Lotka-Volterra} (a) and \textbf{Prokaryotic autoregulator} (b) models, assessed using the log probability of the true generative parameter vector, summarised across the $10$ datasets. The log probabilities were obtained by fitting a mixture of multivariate Gaussian
densities to $500$ samples drawn from an estimate of $p(\thb|\yb)$ obtained using each method. }
\label{Figure:logprob}
\end{figure}
In Figure~\ref{Figure:logprob} we have compared the accuracy of parameters estimates obtained using ABC-SMC and the NLFI methods. Accuracy of the estimates were evaluated as the log probability of the true parameter vector under a mixture of multivariate Gaussian densities fitted to $500$ samples drawn from an estimate of $p(\thb|\yb)$ obtained using each method. We did not notice any drastic difference in accuracy and thus the difference in the estimation of posterior predictive were largely influenced by the estimates of the hidden states. The marginal densities of the parameter posteriors, for one dataset, are shown in Appendix G.2.   

\section{CONCLUSION}

Neural likelihood-free methods have been previously benchmarked using implicit HMMs and are proposed as a computationally cheaper alternative to classical methods such as ABC-SMC in surrounding literature. We have shown that both classical as well as neural likelihood-free methods, by ignoring accurate state estimation, can lead to a grossly incorrect assessment of the goodness-of-fit. We thus proposed a novel technique to approximately estimate the hidden states once samples from the posterior (of the parameters) have been obtained using any likelihood-free method. Our technique, based on learning the posterior Markov process, using an autoregressive flow, produced estimate of the hidden states closer to what can be obtained using SMC, albeit with much fewer simulations.

\subsubsection*{Acknowledgements}

We like to thank the anonymous reviewers for their helpful comments and suggestions. We like to also thank the anonymous reviewers of NeurIPS 2022 and ICLR 2023 conferences, who had kindly provided constructive comments on an earlier version of this work. SG was supported by the Medical Research Council (Unit programme number MC UU 00002/11). 

\bibliographystyle{unsrtnat}
\bibliography{references}

\begin{thebibliography}{44}
\providecommand{\natexlab}[1]{#1}
\providecommand{\url}[1]{\texttt{#1}}
\expandafter\ifx\csname urlstyle\endcsname\relax
  \providecommand{\doi}[1]{doi: #1}\else
  \providecommand{\doi}{doi: \begingroup \urlstyle{rm}\Url}\fi

\bibitem[Sisson et~al.(2018)Sisson, Fan, and Beaumont]{sisson2018handbook}
Scott~A Sisson, Yanan Fan, and Mark Beaumont.
\newblock \emph{Handbook of approximate Bayesian computation}.
\newblock CRC Press, 2018.

\bibitem[Martin et~al.(2019)Martin, McCabe, Frazier, Maneesoonthorn, and Robert]{martin2019auxiliary}
Gael~M Martin, Brendan~PM McCabe, David~T Frazier, Worapree Maneesoonthorn, and Christian~P Robert.
\newblock Auxiliary likelihood-based approximate bayesian computation in state space models.
\newblock \emph{Journal of Computational and Graphical Statistics}, 28\penalty0 (3):\penalty0 508--522, 2019.

\bibitem[Picchini(2014)]{picchini2014inference}
Umberto Picchini.
\newblock Inference for sde models via approximate bayesian computation.
\newblock \emph{Journal of Computational and Graphical Statistics}, 23\penalty0 (4):\penalty0 1080--1100, 2014.

\bibitem[Toni et~al.(2009)Toni, Welch, Strelkowa, Ipsen, and Stumpf]{Toni2009}
T.~Toni, D.~Welch, N.~Strelkowa, A.~Ipsen, and M.~P.H Stumpf.
\newblock {Approximate Bayesian computation scheme for parameter inference and model selection in dynamical systems}.
\newblock \emph{Journal of the Royal Society Interface}, 6\penalty0 (31):\penalty0 187--202, February 2009.

\bibitem[Cranmer et~al.(2020)Cranmer, Brehmer, and Louppe]{cranmer2020frontier}
Kyle Cranmer, Johann Brehmer, and Gilles Louppe.
\newblock The frontier of simulation-based inference.
\newblock \emph{Proceedings of the National Academy of Sciences}, 117\penalty0 (48):\penalty0 30055--30062, 2020.

\bibitem[Lueckmann et~al.(2021)Lueckmann, Boelts, Greenberg, Goncalves, and Macke]{lueckmann2021benchmarking}
Jan-Matthis Lueckmann, Jan Boelts, David Greenberg, Pedro Goncalves, and Jakob Macke.
\newblock Benchmarking simulation-based inference.
\newblock In Arindam Banerjee and Kenji Fukumizu, editors, \emph{Proceedings of The 24th International Conference on Artificial Intelligence and Statistics}, volume 130 of \emph{Proceedings of Machine Learning Research}, pages 343--351. PMLR, 13--15 Apr 2021.

\bibitem[Gordon et~al.(1995)Gordon, Salmond, and Ewing]{gordon1995bayesian}
Neil Gordon, David Salmond, and Craig Ewing.
\newblock Bayesian state estimation for tracking and guidance using the bootstrap filter.
\newblock \emph{Journal of Guidance, Control, and Dynamics}, 18\penalty0 (6):\penalty0 1434--1443, 1995.

\bibitem[Gelman et~al.(1996)Gelman, Meng, and Stern]{gelman1996posterior}
Andrew Gelman, Xiao-Li Meng, and Hal Stern.
\newblock Posterior predictive assessment of model fitness via realized discrepancies.
\newblock \emph{Statistica sinica}, pages 733--760, 1996.

\bibitem[Drovandi et~al.(2016)Drovandi, Pettitt, and McCutchan]{drovandi2016exact}
Christopher~C Drovandi, Anthony~N Pettitt, and Roy~A McCutchan.
\newblock Exact and approximate bayesian inference for low integer-valued time series models with intractable likelihoods.
\newblock \emph{Bayesian Analysis}, 11\penalty0 (2):\penalty0 325--352, 2016.

\bibitem[Andrieu et~al.(2010)Andrieu, Doucet, and Holenstein]{andrieu2010particle}
Christophe Andrieu, Arnaud Doucet, and Roman Holenstein.
\newblock Particle markov chain monte carlo methods.
\newblock \emph{Journal of the Royal Statistical Society: Series B (Statistical Methodology)}, 72\penalty0 (3):\penalty0 269--342, 2010.

\bibitem[Schumacher et~al.(2023)Schumacher, B{\"u}rkner, Voss, K{\"o}the, and Radev]{schumacher2023neural}
Lukas Schumacher, Paul-Christian B{\"u}rkner, Andreas Voss, Ullrich K{\"o}the, and Stefan~T Radev.
\newblock Neural superstatistics for bayesian estimation of dynamic cognitive models.
\newblock \emph{Scientific Reports}, 13\penalty0 (1):\penalty0 13778, 2023.

\bibitem[Ryder et~al.(2021)Ryder, Prangle, Golightly, and Matthews]{ryder2021neural}
Thomas Ryder, Dennis Prangle, Andrew Golightly, and Isaac Matthews.
\newblock The neural moving average model for scalable variational inference of state space models.
\newblock In \emph{Uncertainty in Artificial Intelligence}, pages 12--22. PMLR, 2021.

\bibitem[Papamakarios and Murray(2016)]{papamakarios2016fast}
George Papamakarios and Iain Murray.
\newblock Fast $\varepsilon$-free inference of simulation models with bayesian conditional density estimation.
\newblock \emph{Advances in neural information processing systems}, 29, 2016.

\bibitem[Papamakarios et~al.(2019)Papamakarios, Sterratt, and Murray]{papamakarios2019sequential}
George Papamakarios, David Sterratt, and Iain Murray.
\newblock Sequential neural likelihood: Fast likelihood-free inference with autoregressive flows.
\newblock In \emph{The 22nd International Conference on Artificial Intelligence and Statistics}, pages 837--848. PMLR, 2019.

\bibitem[Bishop(1994)]{bishop1994mixture}
Christopher~M Bishop.
\newblock Mixture density networks.
\newblock 1994.

\bibitem[Rezende and Mohamed(2015)]{rezende2015variational}
Danilo Rezende and Shakir Mohamed.
\newblock Variational inference with normalizing flows.
\newblock In \emph{International conference on machine learning}, pages 1530--1538. PMLR, 2015.

\bibitem[Papamakarios et~al.(2021)Papamakarios, Nalisnick, Rezende, Mohamed, and Lakshminarayanan]{papamakarios2021normalizing}
George Papamakarios, Eric Nalisnick, Danilo~Jimenez Rezende, Shakir Mohamed, and Balaji Lakshminarayanan.
\newblock Normalizing flows for probabilistic modeling and inference.
\newblock \emph{Journal of Machine Learning Research}, 22\penalty0 (57):\penalty0 1--64, 2021.

\bibitem[Parno(2015)]{parno2015transport}
Matthew~David Parno.
\newblock \emph{Transport maps for accelerated Bayesian computation}.
\newblock PhD thesis, Massachusetts Institute of Technology, 2015.

\bibitem[Tabak and Turner(2013)]{tabak2013family}
Esteban~G Tabak and Cristina~V Turner.
\newblock A family of nonparametric density estimation algorithms.
\newblock \emph{Communications on Pure and Applied Mathematics}, 66\penalty0 (2):\penalty0 145--164, 2013.

\bibitem[Cranmer et~al.(2015)Cranmer, Pavez, and Louppe]{cranmer2015approximating}
Kyle Cranmer, Juan Pavez, and Gilles Louppe.
\newblock Approximating likelihood ratios with calibrated discriminative classifiers.
\newblock \emph{arXiv preprint arXiv:1506.02169}, 2015.

\bibitem[Durkan et~al.(2018)Durkan, Papamakarios, and Murray]{durkan2018sequential}
Conor Durkan, George Papamakarios, and Iain Murray.
\newblock Sequential neural methods for likelihood-free inference.
\newblock \emph{arXiv preprint arXiv:1811.08723}, 2018.

\bibitem[Greenberg et~al.(2019)Greenberg, Nonnenmacher, and Macke]{greenberg2019automatic}
David Greenberg, Marcel Nonnenmacher, and Jakob Macke.
\newblock Automatic posterior transformation for likelihood-free inference.
\newblock In \emph{International Conference on Machine Learning}, pages 2404--2414. PMLR, 2019.

\bibitem[Lueckmann et~al.(2017)Lueckmann, Goncalves, Bassetto, {\"O}cal, Nonnenmacher, and Macke]{lueckmann2017flexible}
Jan-Matthis Lueckmann, Pedro~J Goncalves, Giacomo Bassetto, Kaan {\"O}cal, Marcel Nonnenmacher, and Jakob~H Macke.
\newblock Flexible statistical inference for mechanistic models of neural dynamics.
\newblock \emph{Advances in neural information processing systems}, 30, 2017.

\bibitem[Chen et~al.(2021)Chen, Zhang, Gutmann, Courville, and Zhu]{neurSum}
Yanzhi Chen, Dinghuai Zhang, {Michael U.} Gutmann, Aaron Courville, and Zhanxing Zhu.
\newblock Neural approximate sufficient statistics for implicit models.
\newblock In \emph{Ninth International Conference on Learning Representations (ICLR 2021)}, May 2021.
\newblock URL \url{https://iclr.cc/Conferences/2021/Dates}.
\newblock Ninth International Conference on Learning Representations 2021, ICLR 2021 ; Conference date: 04-05-2021 Through 07-05-2021.

\bibitem[Papamakarios et~al.(2017)Papamakarios, Pavlakou, and Murray]{papamakarios2017masked}
George Papamakarios, Theo Pavlakou, and Iain Murray.
\newblock Masked autoregressive flow for density estimation.
\newblock \emph{Advances in neural information processing systems}, 30, 2017.

\bibitem[Germain et~al.(2015)Germain, Gregor, Murray, and Larochelle]{germain2015made}
Mathieu Germain, Karol Gregor, Iain Murray, and Hugo Larochelle.
\newblock Made: Masked autoencoder for distribution estimation.
\newblock In \emph{International Conference on Machine Learning}, pages 881--889. PMLR, 2015.

\bibitem[S{\"a}rkk{\"a}(2013)]{sarkka2013bayesian}
Simo S{\"a}rkk{\"a}.
\newblock \emph{Bayesian filtering and smoothing}.
\newblock Number~3. Cambridge university press, 2013.

\bibitem[Doucet et~al.(2000)Doucet, Godsill, and Andrieu]{doucet2000sequential}
Arnaud Doucet, Simon Godsill, and Christophe Andrieu.
\newblock On sequential monte carlo sampling methods for bayesian filtering.
\newblock \emph{Statistics and computing}, 10\penalty0 (3):\penalty0 197--208, 2000.

\bibitem[Wilkinson(2018)]{wilkinson2018stochastic}
Darren~J Wilkinson.
\newblock \emph{Stochastic modelling for systems biology}.
\newblock CRC press, 2018.

\bibitem[Golightly and Wilkinson(2011)]{golightly2011bayesian}
Andrew Golightly and Darren~J Wilkinson.
\newblock Bayesian parameter inference for stochastic biochemical network models using particle markov chain monte carlo.
\newblock \emph{Interface focus}, 1\penalty0 (6):\penalty0 807--820, 2011.

\bibitem[Durkan et~al.(2020)Durkan, Murray, and Papamakarios]{durkan2020contrastive}
Conor Durkan, Iain Murray, and George Papamakarios.
\newblock On contrastive learning for likelihood-free inference.
\newblock In \emph{International Conference on Machine Learning}, pages 2771--2781. PMLR, 2020.

\bibitem[Hermans et~al.(2020)Hermans, Begy, and Louppe]{hermans2020likelihood}
Joeri Hermans, Volodimir Begy, and Gilles Louppe.
\newblock Likelihood-free mcmc with amortized approximate ratio estimators.
\newblock In \emph{International Conference on Machine Learning}, pages 4239--4248. PMLR, 2020.

\bibitem[Gutmann and Hyv{\"a}rinen(2010)]{gutmann2010noise}
Michael Gutmann and Aapo Hyv{\"a}rinen.
\newblock Noise-contrastive estimation: A new estimation principle for unnormalized statistical models.
\newblock In \emph{Proceedings of the thirteenth international conference on artificial intelligence and statistics}, pages 297--304. JMLR Workshop and Conference Proceedings, 2010.

\bibitem[Tejero-Cantero et~al.(2020)Tejero-Cantero, Boelts, Deistler, Lueckmann, Durkan, Gonçalves, Greenberg, and Macke]{tejero-cantero2020sbi}
Alvaro Tejero-Cantero, Jan Boelts, Michael Deistler, Jan-Matthis Lueckmann, Conor Durkan, Pedro~J. Gonçalves, David~S. Greenberg, and Jakob~H. Macke.
\newblock sbi: A toolkit for simulation-based inference.
\newblock \emph{Journal of Open Source Software}, 5\penalty0 (52):\penalty0 2505, 2020.
\newblock \doi{10.21105/joss.02505}.
\newblock URL \url{https://doi.org/10.21105/joss.02505}.

\bibitem[Gillespie(1977)]{gillespie1977exact}
Daniel~T Gillespie.
\newblock Exact stochastic simulation of coupled chemical reactions.
\newblock \emph{The journal of physical chemistry}, 81\penalty0 (25):\penalty0 2340--2361, 1977.

\bibitem[Pritchard et~al.(1999)Pritchard, Seielstad, Perez-Lezaun, and Feldman]{pritchard1999population}
Jonathan~K Pritchard, Mark~T Seielstad, Anna Perez-Lezaun, and Marcus~W Feldman.
\newblock Population growth of human y chromosomes: a study of y chromosome microsatellites.
\newblock \emph{Molecular biology and evolution}, 16\penalty0 (12):\penalty0 1791--1798, 1999.

\bibitem[Marjoram et~al.(2003)Marjoram, Molitor, Plagnol, and Tavar{\'e}]{marjoram2003markov}
Paul Marjoram, John Molitor, Vincent Plagnol, and Simon Tavar{\'e}.
\newblock Markov chain monte carlo without likelihoods.
\newblock \emph{Proceedings of the National Academy of Sciences}, 100\penalty0 (26):\penalty0 15324--15328, 2003.

\bibitem[Del~Moral et~al.(2012)Del~Moral, Doucet, and Jasra]{del2012adaptive}
Pierre Del~Moral, Arnaud Doucet, and Ajay Jasra.
\newblock An adaptive sequential monte carlo method for approximate bayesian computation.
\newblock \emph{Statistics and computing}, 22\penalty0 (5):\penalty0 1009--1020, 2012.

\bibitem[Marin et~al.(2012)Marin, Pudlo, Robert, and Ryder]{marin2012approximate}
Jean-Michel Marin, Pierre Pudlo, Christian~P Robert, and Robin~J Ryder.
\newblock Approximate bayesian computational methods.
\newblock \emph{Statistics and Computing}, 22\penalty0 (6):\penalty0 1167--1180, 2012.

\bibitem[Kingma and Ba(2015)]{KingmaB14}
Diederik~P. Kingma and Jimmy Ba.
\newblock Adam: {A} method for stochastic optimization.
\newblock In Yoshua Bengio and Yann LeCun, editors, \emph{3rd International Conference on Learning Representations, {ICLR} 2015, San Diego, CA, USA, May 7-9, 2015, Conference Track Proceedings}, 2015.

\bibitem[Golightly and Gillespie(2013)]{golightly2013simulation}
Andrew Golightly and Colin~S Gillespie.
\newblock Simulation of stochastic kinetic models.
\newblock In \emph{In Silico Systems Biology}, pages 169--187. Springer, 2013.

\bibitem[Neal(2003)]{neal2003slice}
Radford~M Neal.
\newblock Slice sampling.
\newblock \emph{The annals of statistics}, 31\penalty0 (3):\penalty0 705--767, 2003.

\bibitem[Filippi et~al.(2013)Filippi, Barnes, Cornebise, and Stumpf]{filippi2013optimality}
Sarah Filippi, Chris~P Barnes, Julien Cornebise, and Michael~PH Stumpf.
\newblock On optimality of kernels for approximate bayesian computation using sequential monte carlo.
\newblock \emph{Statistical applications in genetics and molecular biology}, 12\penalty0 (1):\penalty0 87--107, 2013.

\bibitem[Dean et~al.(2014)Dean, Singh, Jasra, and Peters]{dean2014parameter}
Thomas~A Dean, Sumeetpal~S Singh, Ajay Jasra, and Gareth~W Peters.
\newblock Parameter estimation for hidden markov models with intractable likelihoods.
\newblock \emph{Scandinavian Journal of Statistics}, 41\penalty0 (4):\penalty0 970--987, 2014.

\end{thebibliography}

\end{document}


\onecolumn
\aistatstitle{Sample-efficient neural likelihood-free Bayesian inference of implicit HMMs: \\
Supplementary Materials}

\appendix
\section{Derivations of ABC and incremental posteriors of HMM}

\subsection{Joint distribution for HMM using ABC}
NLFI methods are designed to efficiently sample from the marginal distribution $p(\thb|\yb)$. In ABC although the desired outcome often is the marginal distribution, however it is easy to show that for a latent variable model, such as an implicit HMM, ABC does indeed target an approximation of the joint distribution $p(\thb,\xb|\yb)$.

In ABC we rely upon simulation of a pseudo-data $\hat{\yb}$, when the likelihood $p(\yb|\thb)$ is intractable. The operating principle of any standard ABC algorithm, based on rejection sampling \citep{pritchard1999population}, MCMC \citep{marjoram2003markov} or SMC \citep{Toni2009,del2012adaptive}, is to jointly sample the parameters $\thb$ and the pseudo-data $\ybh$ from their posterior density \citep{marin2012approximate}
\begin{equation}\label{eq:abc pos}
    p_{\epsilon}(\thb,\ybh |\yb) = \frac{\mathbbm{1}_{\epsilon}\left\{d(s(\ybh),s(\yb)<\epsilon)\right\}p(\hat{\yb}|\thb)p(\thb)}{\int \mathbbm{1}_{\epsilon}\left\{d(s(\ybh),s(\yb)<\epsilon)\right\}p(\hat{\yb}|\thb)p(\thb)d\thb},
\end{equation}
where $\mathbbm{1}_{\epsilon}(\cdot)$ is the indicator function, $d(\cdot)$ is a chosen distance metric, $\epsilon>0$ and we consider the summary $s(\cdot)$ to be sufficient. The desired marginal posterior then follows as
\begin{equation}
    p_{\epsilon}(\thb |\yb)=\int p_{\epsilon}(\thb,\ybh |\yb) d\ybh.
\end{equation}
Note that the pseudo-data distribution $p(\hat{\yb}|\thb)$ appearing in \eqref{eq:abc pos} is not required analytically in any of the ABC algorithms. This distribution is essentially the generative model under consideration.

For the HMM such a pseudo data is sampled from the distribution
\begin{equation}
    p(\hat{\yb},\xb|\thb)=\Bigg(\prod_{t=0}^{M-1} g(\ybh_{t}|\Xb_{t},\thb)\Bigg)\Bigg(\prod_{t=1}^{M-1} f(\Xb_{t}|\Xb_{t_1},\thb)\Bigg),
\end{equation}
where $f(\cdot)$, $g(\cdot)$ and thus $p(\hat{\yb},\xb|\thb)$ need not be analytically tractable, just a sample $\ybh$ of the pseudo-data from this distribution is required. Sampling from this distribution is essentially the process of forward sampling from the generative model of the HMM given by \eqref{eq: HMM defn} (see main text). Considering $\ybh$ alone from the pair $(\ybh,\xb)$ we have a sample of the pseudo-data drawn from its marginal $p(\ybh|\thb)$. Thus, when ABC is applied to the HMM in \eqref{eq: HMM defn} the joint density in \eqref{eq:abc pos} is replaced by a density over the triplet $(\thb,\xb, \ybh)$ given by
\begin{equation}\label{eq:abc pos triplet}
    p_{\epsilon}(\thb,\xb, \ybh |\yb) = \frac{\mathbbm{1}_{\epsilon}\left\{d(s(\ybh),s(\yb)<\epsilon)\right\}p(\hat{\yb},\xb|\thb)p(\thb)}{\int \mathbbm{1}_{\epsilon}\left\{d(s(\ybh),s(\yb)<\epsilon)\right\}p(\hat{\yb},\xb|\thb)p(\thb)d\thb},
\end{equation}
from which samples of the pair $(\thb,\xb)$ is distributed from $p_{\epsilon}(\thb,\xb|\yb)$. And the corresponding ABC marginal posterior is given by
\begin{equation}
    p_{\epsilon}(\thb |\yb)=\int p_{\epsilon}(\thb,\xb, \ybh |\yb) d\ybh d\xb .
\end{equation}
From \eqref{eq:abc pos triplet} it is evident that any ABC algorithm applied to the HMM will target the joint distribution $p_{\epsilon}(\thb,\xb|\yb)$. However, this distribution will only be an approximation to the true posterior $p(\thb,\xb|\yb)$, since $\epsilon \ne 0$ (considering $s(\cdot)$ to be sufficient). Note that since $\xb$ is sampled from its prior thus if $\epsilon$ is set to zero (or a small value) then a practically infeasible amount of simulations is required to produce an ABC posterior $p(\thb,\xb|\yb)$ that can approximate closely the true posterior.

\subsection{Deriving the \textit{incremental posterior} decomposition}

We can decompose the posterior of $\xb$, using the product rule, as follows:
\begin{equation}\label{eq:state factor12}
    p(\xb|\thb,\yb) = p(\Xb_{M-1}|\Xb_{M-2:1},\thb,\yb)p(\Xb_{M-2:1}|\thb,\yb).
\end{equation}
Let us first consider the first factor from the above equation, $p(\Xb_{M-1}|\Xb_{M-2:1},\thb,\yb)$. We can obtain from this the density of the last sample points $\Xb_{M-1}$, conditioned on all other random variables, by applying the Markov property and retaining only the terms that involve it, given by:
\begin{equation}\label{eq: factor last}
\begin{aligned}
    p(\Xb_{M-1}|\Xb_{M-2},\ldots, \Xb_1,\thb,\yb) &\propto p(\yb|\Xb_{M-1},\Xb_{M-2},\ldots, \Xb_1,\thb) \\
    & \propto p(\thb)\Bigg(\prod_{t=0}^{M-1} g(\yb_{t}|\Xb_{t},\thb_g)\Bigg)\Bigg(\prod_{t=1}^{M-1} f(\Xb_{t}|\Xb_{t-1},\thb_f)\Bigg)\\
    &\propto
        g(\yb_{M-1}|\Xb_{M-1},\thb_g) f(\Xb_{M-1}|\Xb_{M-2},\thb_f)p(\thb),
\end{aligned}
\end{equation}

which is simply the density $p(\Xb_{M-1}|\Xb_{M-2},\yb_{M-1},\thb)$.

We can also write the conditional distribution of any intermediate sample point $\Xb_t$ among the remaining ones $\Xb_{M-2:1}$, by again applying the Markov property and retaining only the terms that involve it, given by:
\begin{equation}\label{eq: factor intermediate}
\begin{aligned}
p(\Xb_{t}|\Xb_{M-1},\ldots, \Xb_{t+1},\Xb_{t-1},\ldots, \Xb_{1}, \thb,\yb) &\propto p(\yb|\Xb_{M-1},\ldots, \Xb_{t+1},\Xb_{t-1},\ldots, \Xb_{1}, \thb) \\
&\propto p(\thb)\Bigg(\prod_{t=0}^{M-1} g(\yb_{t}|\Xb_{t},\thb_g)\Bigg)\Bigg(\prod_{t=1}^{M-1} f(\Xb_{t}|\Xb_{t-1},\thb_f)\Bigg)\\
&\propto
f(\Xb_{t+1}|\Xb_{t},\thb_f) f(\Xb_{t}|\Xb_{t-1},\thb_f)g(\yb_{t}|\Xb_{t},\thb_g)p(\thb),
\end{aligned}
\end{equation}
which is simply the density $p(\Xb_{t}|\Xb_{t-1},\Xb_{t+1},\yb_{t},\thb)$. 

Using Eq.~\eqref{eq: factor last} and Eq.~\eqref{eq: factor intermediate}, we can now factorise and re-write Eq.~\eqref{eq:state factor12} as given by
\begin{equation}
    p(\xb|\thb,\yb) = p(\Xb_{M-1}|\Xb_{M-2},\thb,\yb)\prod_{t=1}^{M-2} p(\Xb_{t}|\Xb_{t+1}, \Xb_{t-1}, \yb_t, \thb),
\end{equation}
which completes the proof.

\subsection{Pseudocode for the IDE training and prediction}

In Algorithm \ref{alg:IDE train} we provide the pseudocode describing the process of creating a training dataset and then subsequently training the two MAF density estimators emulating the true factor $p(\Xb_{t}|\Xb_{t+1}, \Xb_{t-1}, \yb_t, \thb)$, and the approximate factor $p(\Xb_{t}| \Xb_{t-1}, \yb_t, \thb)$. In Algorithm \ref{alg:IDE pred} we provide the pseudocode for drawing latent sample path (posterior samples of the hidden states) using importance sampling. Note in this case the algorithm expects as input the posterior parameter samples, drawn from the marginal posterior $p(\thb|\yb)$ estimated using NLFI or any other inference method such as ABC.
\begin{algorithm}[H]
   \caption{Simulation and IDE training}
   \label{alg:IDE train}
\begin{algorithmic}
   \STATE {\bfseries Input:} Training dataset size $N$, time series length $M$.
 \STATE  1. Simulate from HMM:
   \FOR{$n=1$ {\bfseries to} $N$}
   \FOR{$t=1$ {\bfseries to} $M-1$}
   \STATE 
   $(\thb^n_f,\thb^n_g, \Xb^n_0)\sim p(\thb), \quad\Xb^n_t \sim f(\Xb_t|\Xb_{t-1},\thb_f),\quad  \yb^n_t \sim g(\yb_t|\Xb_t,\thb_g)$.
   \ENDFOR
   \ENDFOR
   \STATE 2. Generate training examples for the density estimators
   \FOR{$n=1$ {\bfseries to} $N$}
   \FOR{$i=0$ {\bfseries to} $M-3$}
   \FOR{$j=1$ {\bfseries to} $M-2$}
   \FOR{$k=2$ {\bfseries to} $M-1$}
   \STATE  $q_{\phb_{true}}(\Xb_t|\Xb_{t+1},\Xb_{t-1}, \yb_t, \thb)$ emulating the true factor: target  $\Xb^n_j$, inputs $(\Xb^n_k, \Xb^n_i, \yb^n_j, \thb^n)$.
   \STATE $q_{\phb_{appx.}}(\Xb_t|\Xb_{t-1}, \yb_t, \thb)$ emulating the approximate factor: target  $\Xb^n_j$, inputs $(\Xb^n_i,\yb^n_j, \thb^n)$.
   \ENDFOR
   \ENDFOR
   \ENDFOR
   \ENDFOR
\STATE 3. Train the density estimators, using gradient ascent:
\begin{equation}
\begin{aligned}   \phb^*_{true} &=\underset{\phb_{true}}{\operatorname{argmax}}\mathcal{L}(\phb_{true})\\
\phb^*_{appx.} &=\underset{\phb_{appx.}}{\operatorname{argmax}}\mathcal{L}(\phb_{appx.}),
\end{aligned}
\end{equation}
where the loss functions $\mathcal{L}(\phb_{true})$ and $\mathcal{L}(\phb_{appx.})$ are given by the total likelihood of the MAF density estimators:
\begin{equation}
\begin{aligned}
       \mathcal{L}(\phb_{true})&=\sum_{n=1}^N\sum_{i=0,j=1,k=2}^{M-3,M-2,M-1} \log q_{\phb_{true}}(\Xb^n_j|\Xb^n_k, \Xb^n_i, \yb^n_j, \thb^n)\\
       \mathcal{L}(\phb_{appx.})&=\sum_{n=1}^N\sum_{i=0,j=1}^{M-2,M-1} \log q_{\phb_{appx.}}(\Xb^n_j|\Xb^n_i, \yb^n_j, \thb^n).
\end{aligned}
\end{equation}

   \STATE {\bfseries Output:} $\phb^*_{true}, \phb^*_{appx.}$
\end{algorithmic}
\end{algorithm}
\begin{algorithm}[H]
   \caption{Hidden states prediction using IDE}
   \label{alg:IDE pred}
\begin{algorithmic}
   \STATE {\bfseries Input:} Posterior parameter samples $\{\thb^{l}\}_{l=1}^L$ drawn from marginal posterior $p(\thb|\yb)$, time series length $M$, number of importance samples $P$, parameters of trained density estimators $\phb^*_{true}, \phb^*_{appx.}$.\\
 \STATE 1. Generate importance samples
   \FOR{$l=1$ {\bfseries to} $L$}
   
   \FOR{$t=1$ {\bfseries to} $M-1$}
   \FOR{$p=1$ {\bfseries to} $P$}
   \STATE Draw importance samples of the hidden states
   $\Xb^{l,p}_t \sim q_{\phb^*_{appx.}}(\cdot|\Xb^{l,p}_{t-1}, \yb_t, \thb^l)$.
   \STATE Obtain importance weights
   $w^{l,p}_t(\Xb^{l}_t) = 
   \frac{q_{\phb^*_{true.}}(\Xb^{l,p}_{t}|\Xb^{l,p}_{t+1}, \Xb^{l,p}_{t-1}, \yb_t, \thb^l)}{q_{\phb^*_{appx.}}(\Xb^{l,p}_{t}|\Xb^{l,p}_{t-1}, \yb_t, \thb^l)}
   $.
   \ENDFOR
   \ENDFOR
    \ENDFOR
    \STATE 2. Generate weighted samples
    \FOR{$l=1$ {\bfseries to} $L$}
    \FOR{$t=1$ {\bfseries to} $M-1$}
   \FOR{$p=1$ {\bfseries to} $P$}
   \STATE Resample an index $r$ from the set $\{1,\ldots, P\}$, with respective weights $\{w^{l,1}_t,\ldots, w^{l,P}_t \}$.
   \STATE Set $\Xb^{l,p}_{t}=\Xb^{l,r}_{t}$.
      \ENDFOR
   \ENDFOR
    \ENDFOR
   \STATE {\bfseries Output:} $\Xb\in \mathbb{R}^{M \times P \times L}$.
\end{algorithmic}
\end{algorithm}
\section{Nonlinear Gaussian state-space model}
\subsection{Model details}
Here we want to evaluate how well the IDE can perform in comparison to an optimal SMC algorithm which uses the approximate incremental posterior $p(\Xb_t|\Xb_{t-1},\yb_t, \thb)$ as the importance proposal. This density is tractable for Gaussian state-space models. Thus, for this evaluation we have chosen the following state-space model:
\begin{equation}
    \begin{aligned}
        \Xb_t &\sim \mathcal{N}(\bv{A} \gamma(\Xb_{t-1} ), \sigma^2_x \mathbb{I}) \quad t\geq 1\\
        \yb_t &\sim \mathcal{N}(\bv{B} \Xb_t, \sigma^2_y \mathbb{I}),
    \end{aligned}
\end{equation}
where $\gamma(\Xb )=\sin (\exp(\Xb_{t-1}))$, applied elementwise, $\bv{A}=\mathbb{I}_{K\times K}$, $B=2\bv{A}$ and $\Xb_0=\bv{0}$. 

We considered the dimensionality of the state-space, $\operatorname{dim}(\Xb_t)$ and $\operatorname{dim}(\yb_t)$ to be the same, $K=L=10$. We also considered the parameters $\thb=(\sigma_x,\sigma_y)$ to be fixed and known. Thus, we can drop $\thb$ from the conditioning variables for the true and approximate incremental posterior $p(\Xb_t|\Xb_{t-1},\Xb_{t+1},\yb_t)$ and $p(\Xb_t|\Xb_{t-1},\yb_t)$ respectively. And we do the same for the corresponding density estimates: $q_{\phb}(\Xb_t|\Xb_{t-1}, \yb_t)$ and $q_{\phb}(\Xb_t|\Xb_{t-1}, \Xb_{t+1},\yb_t)$. For the model above, the approximate incremental posterior is known analytically and happens to be a Gaussian:
\begin{equation}
    p(\Xb_t|\Xb_{t-1},\yb_t) = \mathcal{N}(\Xb_t;\bv{m}, \bv{\Sigma}),
\end{equation}
whose mean and the covariance are given by
\begin{equation}
    \begin{aligned}
        \bv{\Sigma}^{-1} &= \Sigma^{-1}_x + B\Sigma^{-1}_y B\\
        \bv{m} &= \bv{\Sigma}(\Sigma^{-1}_x \gamma(\Xb_{t-1})  +  B\Sigma^{-1}_y \yb_t),
    \end{aligned}
\end{equation}
where $\Sigma_x=\sigma^2_x \mathbb{I}$ and  $\Sigma_y=\sigma^2_y \mathbb{I}$. 

We used $\sigma_x=\sigma_y=0.5$ to generate the simulated data. We considered a long time series with $M=500$ time points. We created the IDE training set as was described in section 4.2 (main text).

For the IDE's MAF we have used $J=3$ transformations, each of which has two hidden layers of $50$ units and ReLU nonlinearities. We found that chaining a few transformations was enough to learn a Gaussian density. Increasing the number of transformations did not improve the performance noticeably. For training the MAF we used ADAM \citep{KingmaB14} with a minibatch size of $256$, and a learning rate of $0.0005$. Following, \cite{papamakarios2019sequential} we used $10\%$ of the training data as a validation set, and stopped training if validation log likelihood did not improve after $20$ epochs.

\subsection{Additional experiments with state-space model}

\begin{figure}[!ht]

  \centering
  
  \begin{subfigure}[]{
    \includegraphics[width=.7\textwidth]{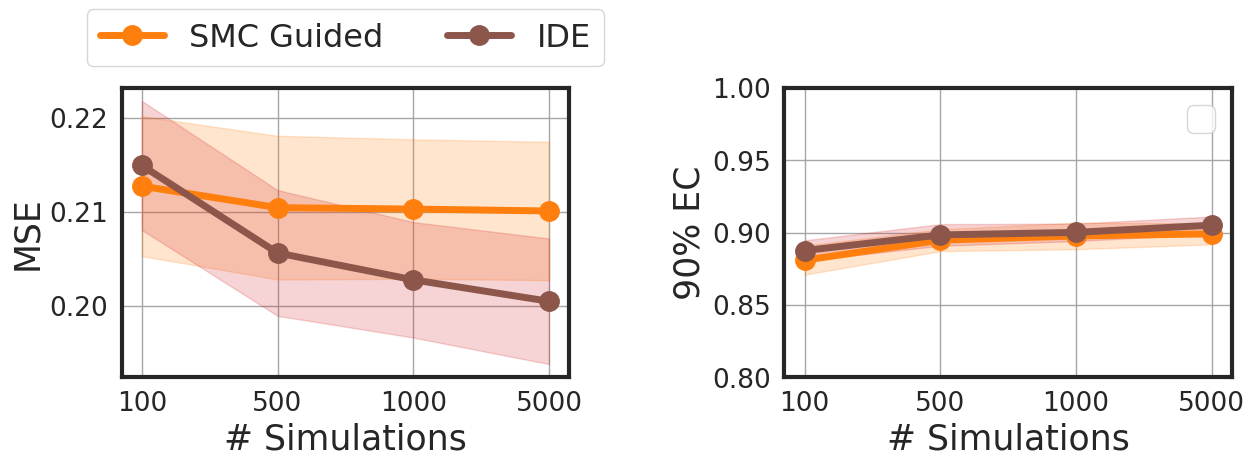}
    \label{Figure:state1}
  }  
  \end{subfigure}
    \begin{subfigure}[]{
    \includegraphics[width=.7\textwidth]{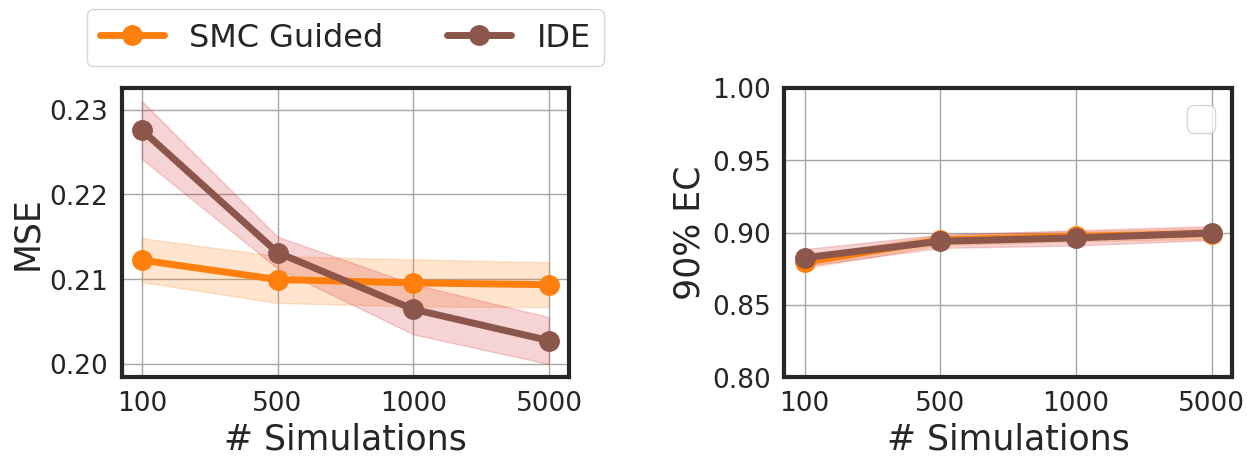}
    \label{Figure:state2}
}\end{subfigure}

\caption{Estimation of the \textbf{hidden states} of a \textbf{nonlinear state-space} model, for two different experiments: (a) $\sigma_x=\sigma_y=1$ and $K=10$, (b) $\sigma_x=\sigma_y=1$ and $K=30$. The quality of approximations was quantified using the MSE and $90 \%$ EC, summarised using the mean (solid line) and $95 \%$ confidence intervals (shaded area), across $10$ datasets. }
\label{Figure:state}
\end{figure}

In the main text we have presented results for using parameters $\sigma_x=\sigma_y=0.5$. However, we have carried out additional experiments firstly with noise $\sigma_x=\sigma_y=1$ and then probing the performances for even more higher-dimensional states space, $K=30$, along with this higher noise setting. See Figure \ref{Figure:state} for the results of these additional experiments. Note that we consistently found the Bootstrap SMC to give extremely poor performance, and thus not shown in the plots.

\section{Model details}

\subsection{Stochastic Lotka-Volterra model}

The stochastic Lotka-Volterra model, a stochastic kinetic system, can be defined through the following list of reactions:
\begin{equation}
\begin{aligned}
&\mathcal{R}_1: \quad X^{prey} \overset{c_1}{\longrightarrow}2 X^{prey}\\
&\mathcal{R}_2: \quad X^{prey} + X^{pred} \overset{c_2}{\longrightarrow}2 X^{pred}\\
&\mathcal{R}_3: \quad X^{pred} \overset{c_2}{\longrightarrow} \emptyset,
\end{aligned}
\end{equation}
where we denote by $X^{prey},X^{pred}$ the prey and predator species respectively. We further denote the corresponding numbers of the species as the system state $\Xb_t=(X^{prey}_t,X^{pred}_t)$. The hazard vector for this system is $h(\Xb_t,\cb)=\big(c_1 X^{prey}_t,c_2 X^{prey}_t X^{pred}_t,c_3 X^{pred}_t\big)$. The stoichiometry matrix for this system is given by
\begin{equation}
S= 
\begin{pmatrix}
1 & -1 & 0\\
0 & 1 & -1
\end{pmatrix}.
\end{equation}
We set the initial values as $\Xs_0=(100,100)$ and consider them known. 

A MJP describing a stochastic kinetic system, like the one above or the PKY model, is characterised by the transition probability $p(t_0,\Xb_0,t,\Xb_t):=p(\Xb,t)$ for the process arriving at state $\Xb_t$ at time $t$ conditioned on an initial state $\Xb_0$ at time $t_0$. This is basically the density $f(\cdot)$ in equation \eqref{eq: HMM defn} in main text, in continuous time. Now this transition probability is given by the solution of the following differential equation:
\begin{equation}
    \frac{\partial p(\Xb,t)}{\partial t}=\sum_{i=1}^v = \{h_i(\Xb-S^i,c_i)p(\Xb-S^i,t) - h_i(\Xb,c_i)p(\Xb,t)\},
\end{equation}
known as the chemical master equation \citep[and the references therein]{golightly2013simulation}. The CME only admits an analytical solution for a handful of simple models (not for the ones we have used: LV and PKY). Thus, the density $f(\cdot)$ cannot be evaluated. However, the seminal work in \cite{gillespie1977exact} developed an algorithm, commonly referred to as the \textit{stochastic simulation algorithm}, that can simulate $\Xb$ exactly.

We generated simulated trajectories from this model using the stochastic simulation algorithm and added Gaussian noise corruption, with variance $100$, at $50$ time points. We used the following generative values of the parameters $\thb=(0.3,0.0025,0.5)$ to ensure that the model follows an oscillatory regime. Moreover, following previous studies we considered the initial values to be known and set at $\Xb_{t_0}=(100,100)$. 

We used the following set of \textbf{prior distributions}: $c_1\sim\operatorname{Beta}(1,2)$, $c_2\times 10^3\sim\mathcal{U}(15,50)$ and $c_3\sim\operatorname{Beta}(2,1)$.

For running ABC-SMC and all the NLFI methods we downsampled the generated time series by a factor of $5$ to create a \textbf{summary statistic} $s(\yb)\in \mathbb{R}^{20}$ which is used in place of the full data $\yb$. 

\subsection{Prokaryotic autoregulatory gene network}

We considered the autoregulatory model used to benchmark the particle MCMC method in \cite{golightly2011bayesian}. This is a simplified model that describes a mechanism for autoregulation in prokaryotes based on a negative feedback mechanism of dimers of a protein coded by a gene repressing its own transcription. Essentially this is a stochastic kinetic model described by the following set of reactions:
\begin{equation}
    \begin{aligned}
           &\mathcal{R}_1: DNA + P2 \rightarrow DNA\cdot P2 \\
           &\mathcal{R}_2: DNA \cdot P2 \rightarrow DNA + P2 \\
&\mathcal{R}_3: DNA \rightarrow DNA + RNA \\
&\mathcal{R}_4:  RNA \rightarrow RNA + P\\
&\mathcal{R}_5: 2P \rightarrow P2 \\
&\mathcal{R}_6: P2 \rightarrow 2P\\
& \mathcal{R}_7: RNA \rightarrow \emptyset\\
& \mathcal{R}_8: P \rightarrow \emptyset.
    \end{aligned}
\end{equation}
We order the variables as $\Xb= (RNA,P,P2,DNA,DNA\cdot P2)$ leading to a stoichiometry matrix for the system:
\begin{equation}
    S = \begin{pmatrix}
    0 & 0 & 1 & 0 & 0 & 0 & -1 & 0\\
    0 & 0 & 1 & -2 & 2 & 0 & -1\\
    -1 & 1 & 0 & 0 & 1 & -1 & 0 & 0\\
    1 & -1 & 0 & 0 & 0 & 0 & 0 & 0\\
    -1 & 1 & 0 & 0 & 0 & 0 & 0 & 0
    \end{pmatrix},
\end{equation}
and the associated hazard function is given by
\begin{equation}
    h(\Xb, \cb) = (c_1 DNA \times P2, c_2 DNA \cdot P2, c_3 DNA, c_4 RNA, c_5 P(P -1)/2, c_6 P2, c_7 RNA, c_8 P).
\end{equation}
This model has one conservation law \citep{golightly2011bayesian}
\begin{equation}
    DNA \cdot P2 + DNA = k,
\end{equation}
where $k$ is the number of copies of this gene in the genome. Following \cite{golightly2011bayesian} we use this relation to to remove
$DNA \cdot P2 $ from the model, replacing any occurrences of $DNA \cdot P2 $ in rate laws with $k - DNA$. This leads to a reduced full-rank model with species $\Xb = (RNA, P, P2, DNA)$, stoichiometry matrix:
\begin{equation}
    S = \begin{pmatrix}
    0 & 0 & 1 & 0 & 0 & 0 & -1 & 0\\
    0 & 0 & 1 & -2 & 2 &  & -1\\
    -1 &1 &0 &0 &1 &-1 &0 &0\\
    -1 &1 &0 &0 &0 &0 &0 &0
    \end{pmatrix},
\end{equation}
and associated hazard function
\begin{equation}
    h(\Xb, \cb) = (c_1 DNA \times P2, c_2 (k - DNA), c_3 DNA, c_4 RNA, c_5 P(P -1)/2, c_6 P2, c_7 RNA, c_8 P).
\end{equation}
We consider $k$ to be known and set to $10$. Again we generated simulated trajectories from this model using the stochastic simulation algorithm.

Following \cite{golightly2011bayesian}, we considered the observations as a linear combination of the proteins $P,P2$ as follows:
\begin{equation}
    y_t = P_t + 2P2_t + \epsilon_t,
\end{equation}
where $\epsilon$ is assumed to be iid Gaussian noise. We generated $100$ simulated observations from this model at times $t=[0:.5:50]$ with generative rate constants $\thb=(0.1,0.7,0.35,0.2,0.1,0.9,0.3,0.1)$ and $\epsilon \sim \mathcal{N}(0,4)$. In this case also we considered the initial values $\Xb_{t_0}$ to be known and set to $(8,8,8,5)$.  

We placed a $\operatorname{Gamma}(2,3)$ \textbf{prior} on all the rate constants.

We downsampled the simulated data by a factor of five to obtain the \textbf{summary statistic} $s(\yb)\in\mathbb{R}^{20}$.

\section{NLFI, IDE and ABC-SMC implementation details for biological HMMs}\label{implementations}
For SNLE we used a MAF as the likelihood density estimator $q_{\psi}(s(\yb)|\thb)$ and for SRE we used a MLP classifier. For both uses of the MAFs, $q_{\phb}(\Xb_t|\Xb_{t-1}, \yb_t, \thb)$ and $q_{\phb}(\Xb_t|\Xb_{t+1}, \Xb_{t-1}, \yb_t, \thb)$ for the IDE and $q_{\psi}(\thb|s(\yb))$ for SNLE, we used the same architecture. That is $J=5$ transformations each of which
has two hidden layers of $50$ units each and ReLU nonlinearities. For SRE we used a residual network based classifier with two residual layers of $50$ units each and ReLU nonlinearities. 

For training all the neural networks we used ADAM \citep{KingmaB14} with the same minibatch size, learning rate and validation split as was used for the experiment with the state-space model. Following \cite{papamakarios2019sequential}, we used the Slice Sampling algorithm \citep{neal2003slice} to draw samples from the posterior while using SNLE and SRE.

We applied the particular version of ABC-SMC algorithm, that was proposed in \cite{Toni2009}, using $1000$ particles. Furthermore, we used an adaptive tolerance sequence where the tolerance $\epsilon_{\tau}$ at the $\tau$-th step of the algorithm is selected as the $0.1$-quantile of the distances of the accepted particles in the $\tau-1$-th step. Moreover, we chose the perturbation kernel of ABC-SMC (see \cite{Toni2009}) as a multivariate Gaussian whose covariance is based on a \textit{k-nearest neighbours} strategy, with $k=15$, proposed in \cite{filippi2013optimality}. We terminated the ABC-SMC algorithm when a predetermined number of simulations has been carried out. If that number is exceeded within the $\tau$-th step, we then considered the weighted particle system at the $\tau-1$-th step as the desired ABC posterior. 

\section{Evaluations without using summary statistics}
\begin{figure*}

  \centering
  
  \begin{subfigure}[]{
    \includegraphics[width=0.89\textwidth,height=.15\textheight]{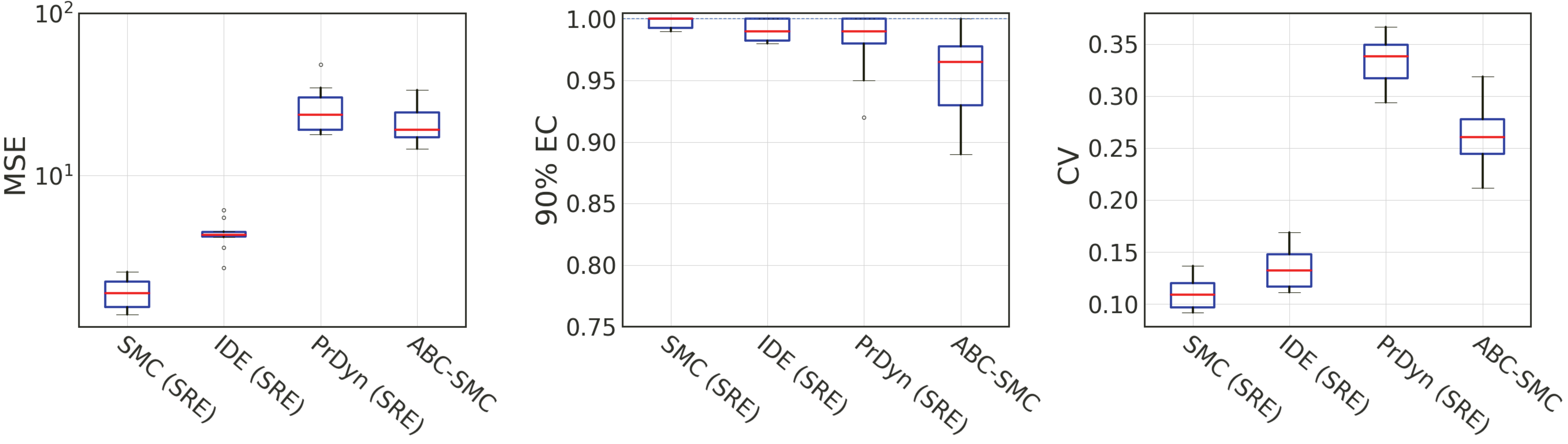}
    \label{Figure:ppd ns}
  }  
  \end{subfigure}
	\begin{subfigure}[]{
    
    \includegraphics[width=0.89\textwidth,height=.15\textheight]{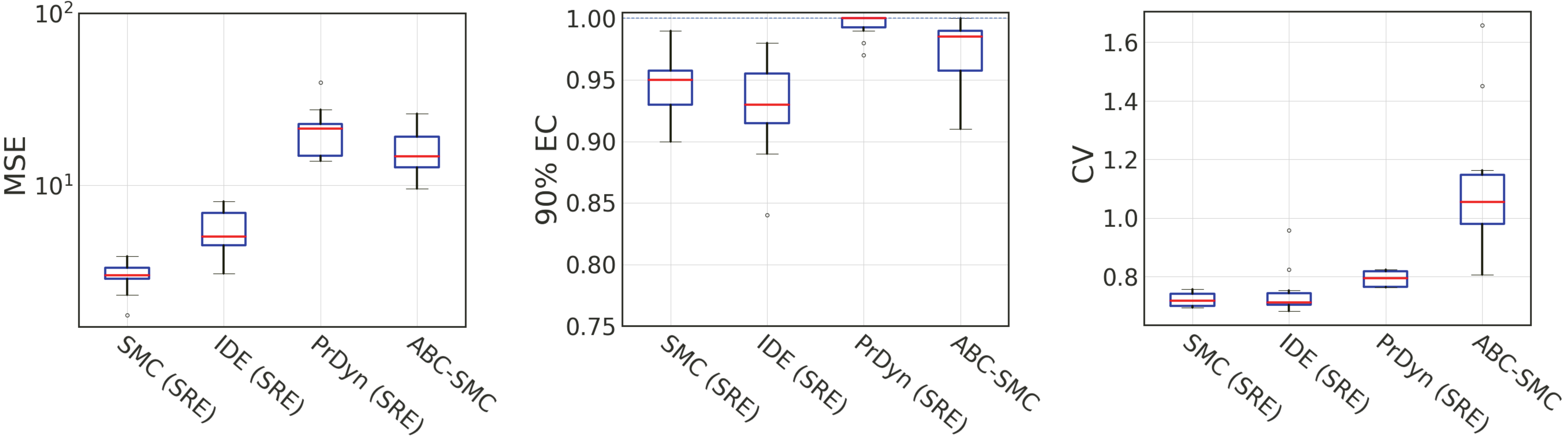}
    \label{Figure: states ns}
}\end{subfigure}

\caption{Comparison of the estimates of the (a) \textbf{posterior predictive distribution} and (b) \textbf{hidden states} of the \textbf{Prokaryotic autoregulator} models. We summarised the chosen metrics across $10$ simulated datasets. The baseline is \textbf{SMC}. Here we are using the full data rather than the summaries.}
\label{Figure:ppd ns}
\end{figure*}
All our evaluations on the two biological HMMs were based on the use of hand-crafted summary statistics. Here we repeat the analysis for the PKY model without using summary statistics. For ABC-SMC this means calculating a distance between the full observed data (considering all the time points) and the simulated one. Note that the particular ABC-SMC algorithm that we have used \citep{Toni2009} was originally designed to work with full data. For obtaining the hidden states 
 and subsequently the posterior predictive distribution using SMC, IDE and PrDyn we have used an estimate of $\thb$ obtained using SRE trained on the full dataset. For this we extended the classifier neural network with a $2$-layer LSTM, trained simultaneously with the classifier, to embed the data into a smaller dimensional summary statistics. We used a LSTM with a $10$-dimensional hidden state and fed the hidden state, corresponding to the last time-step, into a fully connected layer consisting $8$ hidden units and a ReLU activation function. Thus, we have a $8$-dimensional summary statistics that is learnt on the fly.
\begin{figure*}[h]
\centering
\includegraphics[width=0.9\textwidth,height=0.3\textheight]{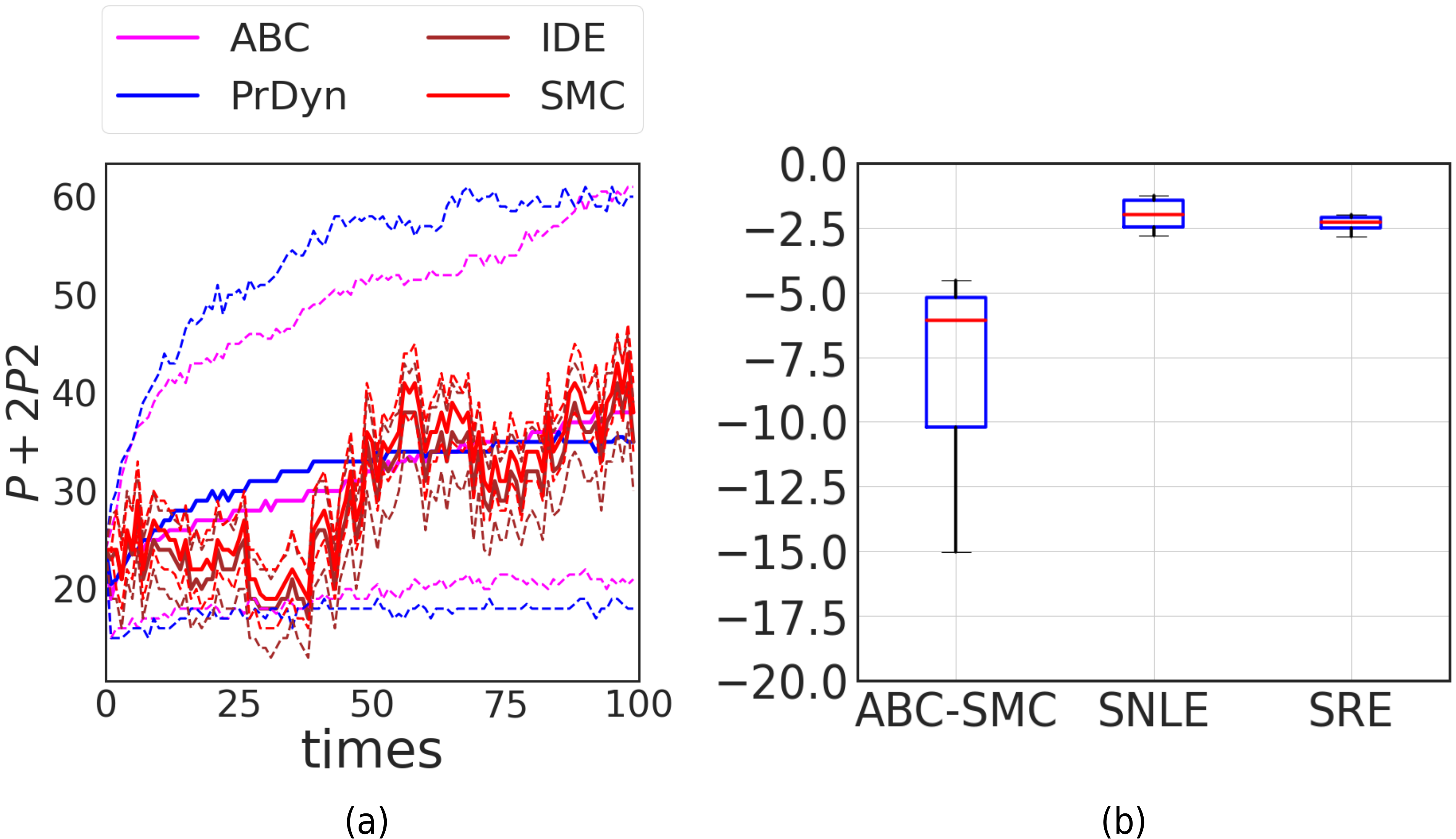}
    
\caption{(a) Posterior distributions of the latent sample path $\xb$ summarised by the mean (solid lines) and $95\%$ credible intervals (broken lines), for the \textbf{Prokaryotic autoregulator}. The \textbf{ABC-SMC} is using the full dataset. (b) Accuracy of parameter estimates for the \textbf{Prokaryotic autoregulator} model, evaluated using the log probability of the true generative parameter vector, summarised across the $10$ datasets. \textbf{SRE} and \textbf{ABC-SMC} is using the full dataset.}
\label{Figure:pkynosum}
\end{figure*}

 In Figure~\ref{Figure:ppd ns} we compare the estimates of the posterior predictive and the hidden states using the same metrics that we have used previously. We noticed that the IDE produced estimates of these quantities closer to the baseline (SMC's estimate) than ABC-SMC and PrDyn. Additionally, we noticed a slight improvement of ABC-SMC's performance in estimating the hidden states (see also Figure~\ref{Figure:pkynosum} (a) where we have plotted the estimated hidden states for one dataset), however the accuracy of the parameters estimates (summarised in Figure~\ref{Figure:pkynosum} (b)) did not change significantly from what was observed while using summary statistics. Note that the accuracy of the parameter estimates did not change significantly for the SRE as well. Despite having access to the full data the ABC-SMC's proposal mechanism for the hidden states is still too inefficient to significantly improve the accuracy of reconstructing the hidden states within a practically feasible simulation budget. 

 \section{Joint inference of the sample path and parameters using a MAF}

We have argued before (see the last paragraph of section 3 in main text) that NLFI methods cannot be used directly for inferring the joint posterior $p(\xb,\thb|\yb)$. Next, we have shown results for an experiment, using the LV model, that supports our argument. Note that since we cannot evaluate the joint density $p(\xb,\thb)$, the only strategy that can be applied is of using a normalizing-flow to directly emulate the joint posterior $p(\xb,\thb|\yb)\approx q_{\psb}(\xb,\thb|\yb)$. We denote this approach as neural posterior estimation (NPE). We used $10^6$ simulations from the model to train a MAF representing $q_{\psb}(\xb,\thb|\yb)$. Note that for the proposed IDE approach we used $35 \times 10^3$ (including inference of $\thb$). We retained the same architecture and optimisation settings that we used in other experiments. Once trained, we used one of the simulated dataset for the LV model to carry out inference. This is the same dataset corresponding to the plot shown in Figure~\ref{Figure:sde}. 

\begin{figure*}[h]

  \centering
  
  \begin{subfigure}[]{
    \includegraphics[width=.45\textwidth]{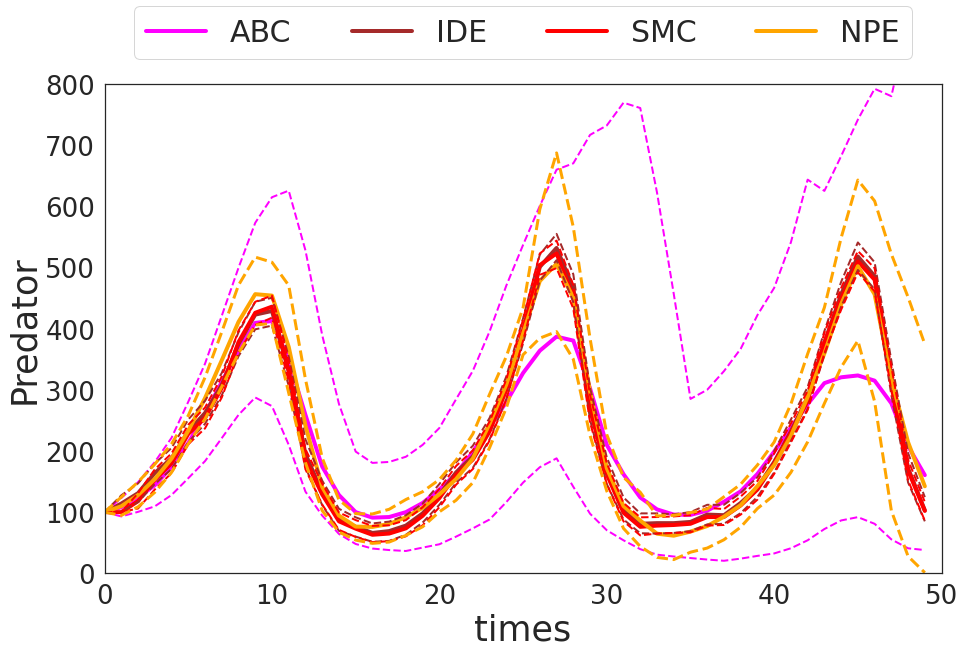}
    \label{Figure:lvlogprob}
  }  
  \end{subfigure}
	\begin{subfigure}[]{
    
    \includegraphics[width=0.45\textwidth]{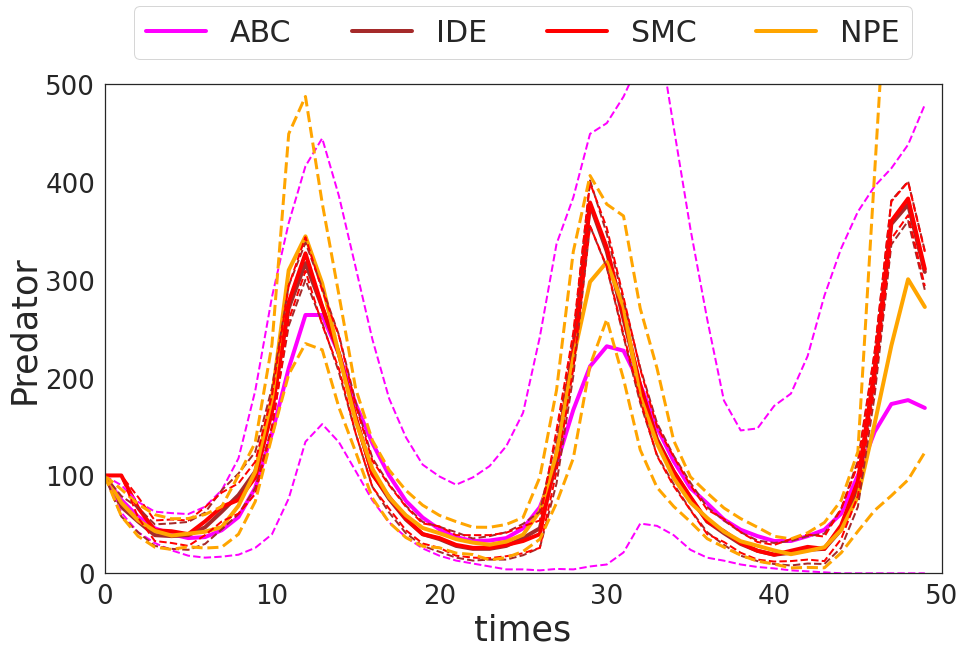}
    \label{Figure:pkylogprob}
}\end{subfigure}
    
\caption{Comparison between methods that estimate jointly the parameters and hidden states of a HMM (in this case the \textbf{Lotka-Volterra} model), such as  \textbf{ABC-SMC} \& \textbf{NPE}, with those that estimate these quantities separately, such as \textbf{SMC} \& \textbf{IDE}. The plot above shows the posteriors of the hidden states summarised by the mean (solid lines) and $95\%$ credible intervals (broken lines). The proposed method \textbf{IDE} reduces the simulation burden by a large factor in comparison to \textbf{NPE}. Note that even with a much larger simulation budget \textbf{NPE} fails to correctly estimate the hidden states as well as the parameters (see Figure~\ref{Figure:lv_pars_npe}).
}
\label{Figure:lv_npe}
\end{figure*}

\begin{figure*}[!ht]
\includegraphics[width=\textwidth,keepaspectratio=true]{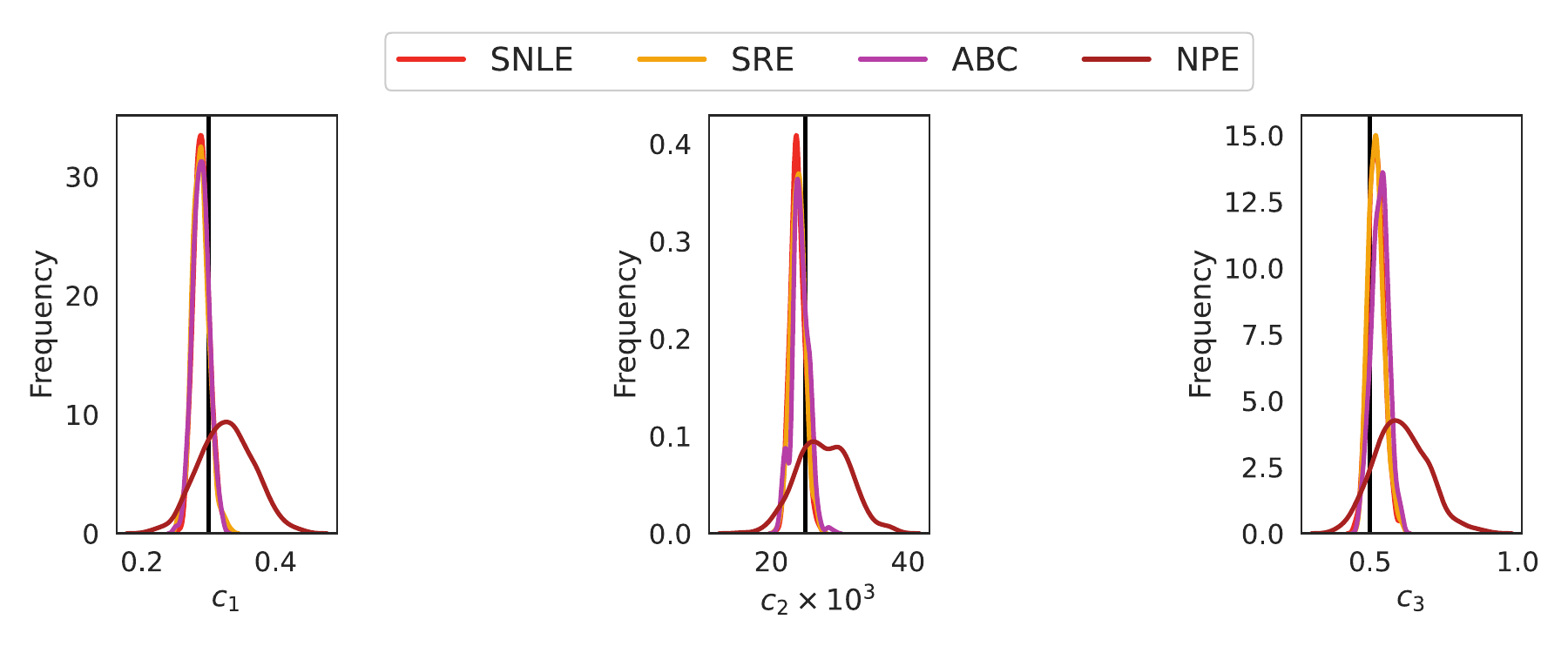}
    
\caption{Posterior marginal densities of the parameters of the \textbf{Lotka-Volterra} model obtained using SNLE, SRE (both targeting the marginal $p(\thb|\yb)$) with NPE, ABC-SMC (both targeting the joint $p(\xb,\thb|\yb)$). NPE failed to estimate $\thb$ correctly. }
\label{Figure:lv_pars_npe}
\end{figure*}

In Figure~\ref{Figure:lv_npe} we plot components of the hidden state estimated by SMC, IDE, ABC-SMC and NPE. Note that SMC, IDE are using same samples of $\thb$ estimated using SNLE. All methods use $500$ samples from the posteriors of $\thb,\xb$. In Figure~\ref{Figure:lv_pars_npe} we show the corresponding parameter estimates. Although NPE estimates the hidden state better than ABC-SMC, its estimation quality drops at those time points where the concentration reaches a peak before decreasing again. This drop is much more pronounced near the last peak. The parameter estimates are however significantly different than all the other methods. From which it can be concluded that NPE performs worse than even ABC-SMC to produce the posterior of the parameters when targeting $\xb,\thb$ jointly.

Additionally, as further pilot experiments, we have also repeated this experiment without using summary statistics for NPE and rather (i) learning the summaries using a LSTM and (ii) feeding in the full data as the input to the normalising-flow. However, the results were even worse and thus we have not shown them here.

\section{Plots of hidden states and parameter posteriors}

\subsection{Plots of hidden states}
\begin{figure}[!ht]

  \centering
  
  \begin{subfigure}[]{
    \includegraphics[width=.465\textwidth]{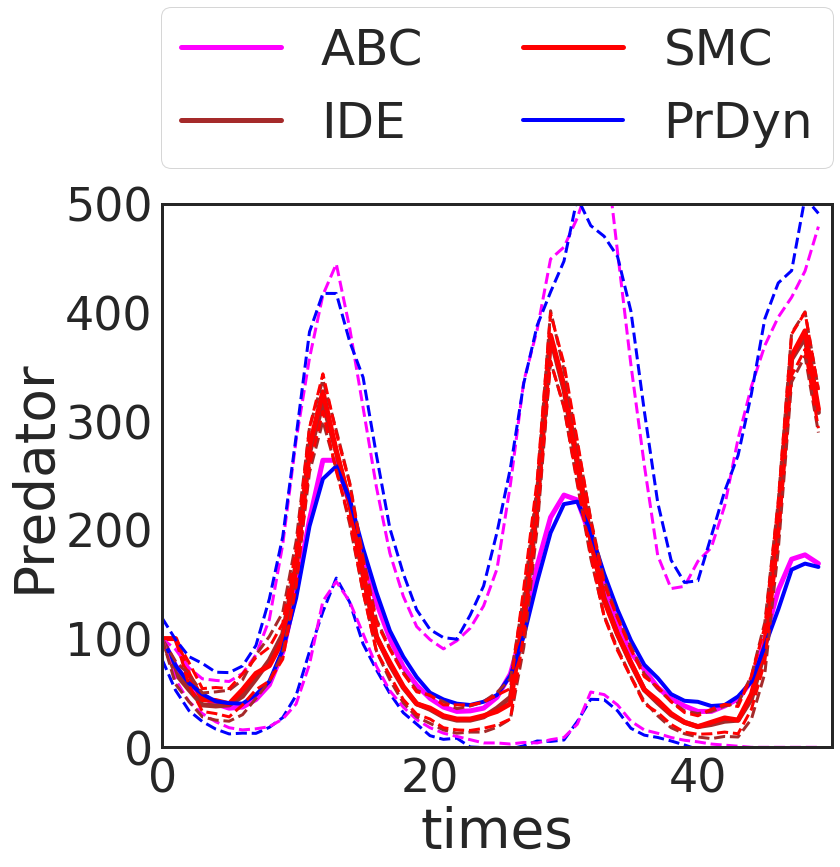}
    \label{Figure:ldfsdprob}
  }  
  \end{subfigure}
	\begin{subfigure}[]{
    
    \includegraphics[width=0.454\textwidth]{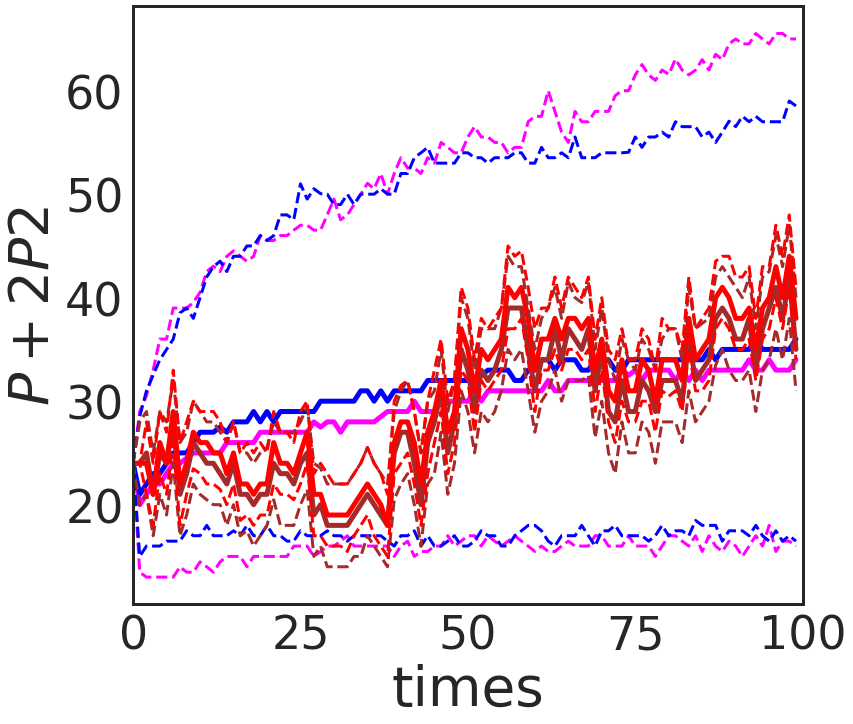}
    \label{Figure:pkysdfsdrob}
}\end{subfigure}

\caption{Posterior distributions of one component of the latent sample path (the hidden states) $\xb$ summarised by the mean (solid lines) and $95\%$ credible intervals (broken lines), for the \textbf{Lotka-Volterra} (a), \textbf{Prokaryotic autoregulator} (b) model. Here SMC, IDE and PrDyn estimates of $\xb$ corresponds to an SNLE estimate of $\thb$.}
\label{Figure:sde}
\end{figure}

The following plots of the posterior sample paths (posterior of the hidden states) for one dataset (Figure~\ref{Figure:sde}), clearly show the overestimation of uncertainty in case of PrDyn and ABC-SMC, for all models. 

\subsection{Plots of marginal posteriors of the parameters}

\begin{figure*}[!ht]
\includegraphics[width=0.8\textwidth,keepaspectratio=true]{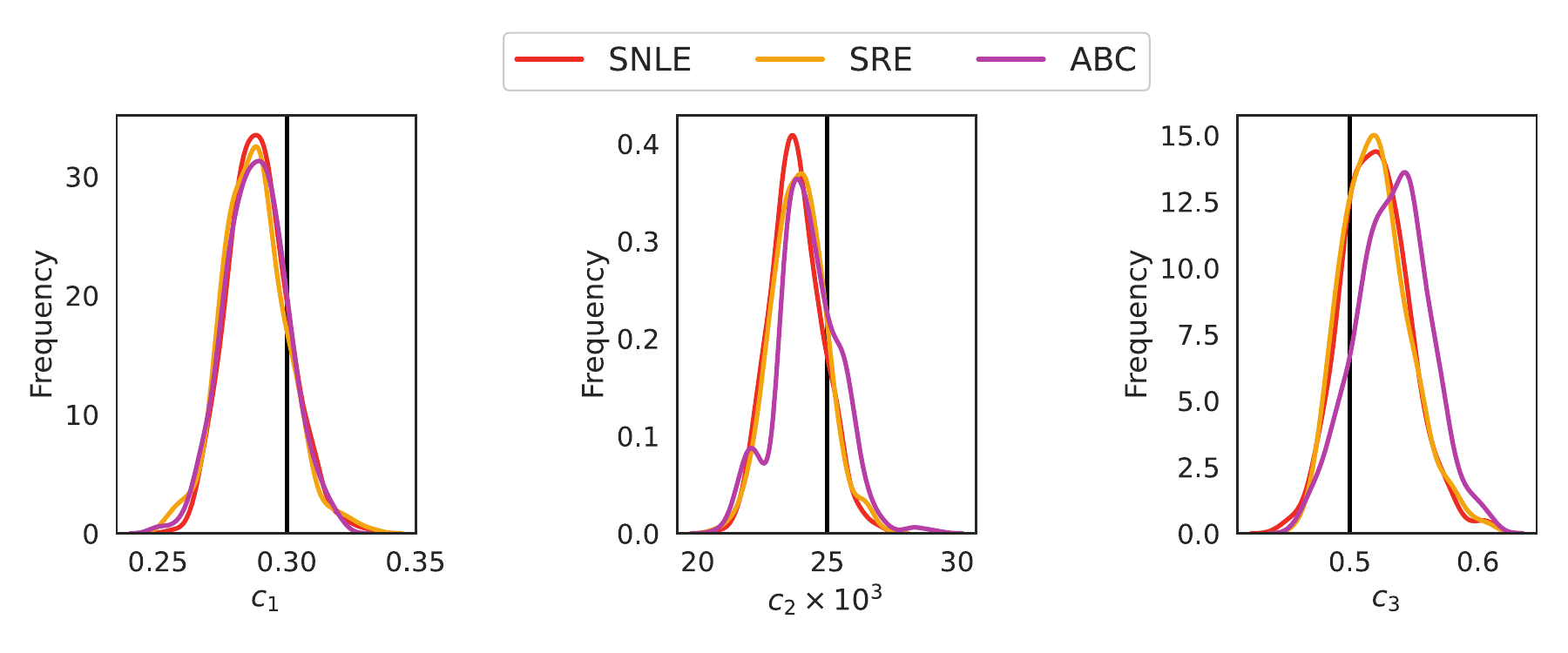}
   
\caption{Posterior marginal densities of the parameters of the \textbf{Lotka-Volterra} model, inferred from one of the $10$ datasets.}
 \label{Figure:lv_pars}
\end{figure*}

\begin{figure*}[!ht]
\includegraphics[width=0.8\textwidth,keepaspectratio=true]{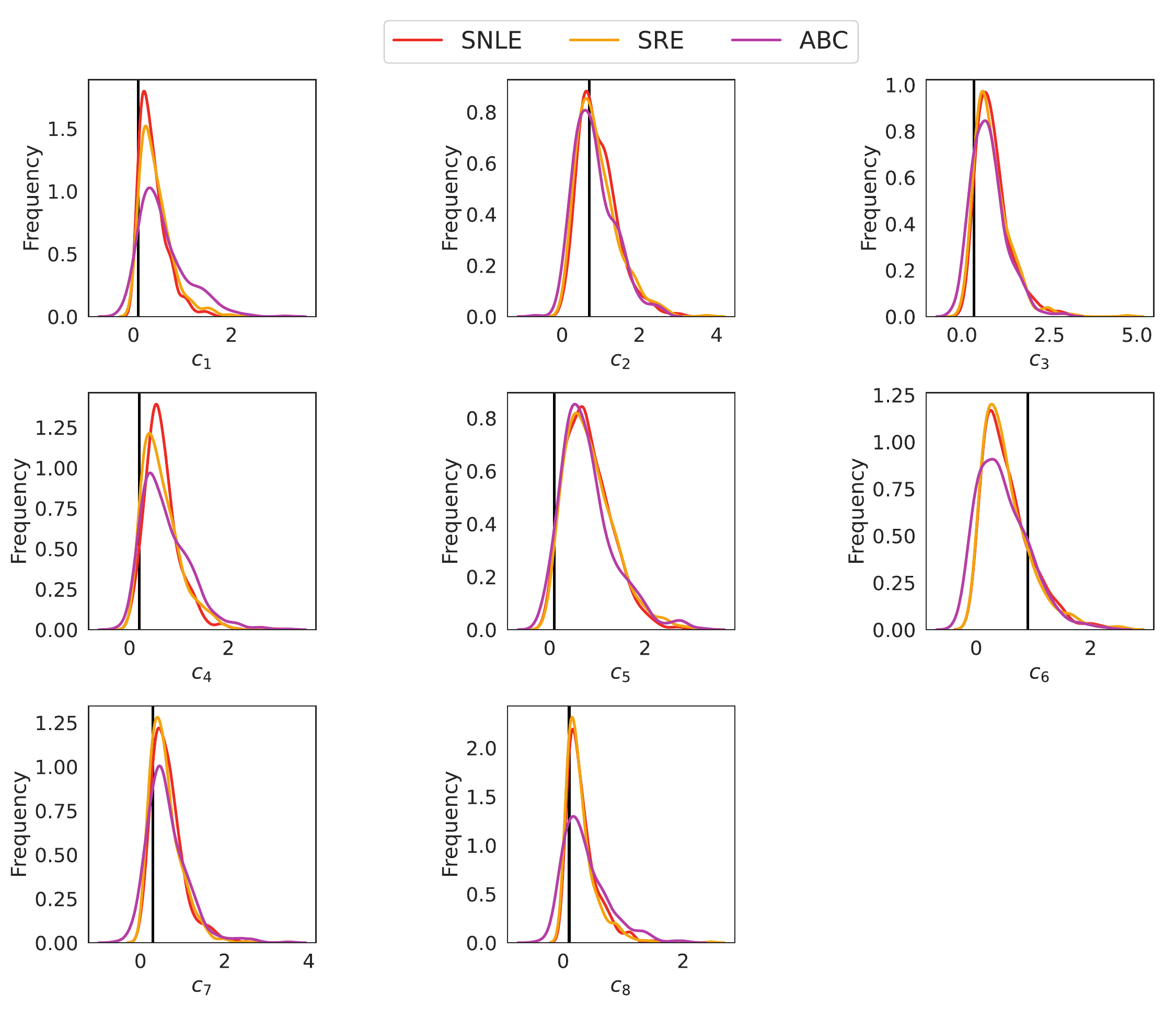}
   
\caption{Posterior marginal densities of the parameters of the \textbf{Prokaryotic autoregulatory} model, inferred from one of the $10$ datasets.}
 \label{Figure:pky_pars}
\end{figure*}

In the subsequent plots Figure~\ref{Figure:lv_pars} and ~\ref{Figure:pky_pars} we compare the parameter estimates of the models between NLFI based methods, SNLE/SRE, and ABC-SMC. Here we have shown the estimates for one of the $10$ different simulated datasets. This is the same dataset corresponding to the plot shown in Figure~\ref{Figure:sde}. Note that the parameter estimates are reasonably close to each other and thus the estimate of the posterior predictive distribution is largely influenced by the estimates of the hidden states.

\section{Related work in inference of implicit HMMs}

The most common approaches to tackle the inference of an implicit HMM consist largely of ABC methods \citep{dean2014parameter,martin2019auxiliary,picchini2014inference}. Note that when the observational density is known analytically then the particle-MCMC \citep{andrieu2010particle} method can be used to carry out exact inference. However, the computational cost of this method is prohibitive, as in each step of MCMC a particle filter with a large number of particles is run to calculate an unbiased estimate of the marginal likelihood. Interestingly, a new avenue of research can be of combining our proposed IDE as an importance density within a particle-MCMC scheme.  An alternative approach which combines SMC with ABC was proposed in \citep{drovandi2016exact}. However, this approach requires the problematic choices of ABC algorithmic parameters.
\clearpage
\bibliography{references.bib}